\documentclass[preprint,review,5p,times,twocolumn]{elsarticle}

\usepackage{graphicx} % provides the includegraphics command
\graphicspath{ {./images/} }
\usepackage{amssymb} % provides various useful mathematical symbols
\usepackage{amsthm} % provides extended theorem environments
\usepackage{float} % for adjustments of figures and tables
\usepackage{lineno} % adds line numbers
\usepackage{soul,color}
\usepackage{caption}
\usepackage{subcaption}
\usepackage{multirow}
\usepackage{amsmath}
\usepackage{hyperref}
\usepackage{lineno}
\usepackage{makecell}
\usepackage{booktabs}

\begin{document}

\begin{frontmatter}

%% Title, authors and addresses

%% use the tnoteref command within \title for footnotes;
%% use the tnotetext command for theassociated footnote;
%% use the fnref command within \author or \address for footnotes;
%% use the fntext command for theassociated footnote;
%% use the corref command within \author for corresponding author footnotes;
%% use the cortext command for theassociated footnote;
%% use the ead command for the email address,
%% and the form \ead[url] for the home page:
%% \title{Title\tnoteref{label1}}
%% \tnotetext[label1]{}
%% \author{Name\corref{cor1}\fnref{label2}}
%% \ead{email address}
%% \ead[url]{home page}
%% \fntext[label2]{}
%% \cortext[cor1]{}
%% \affiliation{organization={},
%%             addressline={},
%%             city={},
%%             postcode={},
%%             state={},
%%             country={}}
%% \fntext[label3]{}

%\title{Raw Data Is All You Need: Virtual Axle Detector with Enhanced Receptive Field}
%\title{Receptive Field Rule: Object Size Driven Design of Convolutional Neural Networks}
%\title{Object Size Driven Design of Convolutional Neural Networks validated on Virtual Axle Detection}
%Leveraging Object Size Driven Design of Convolutinal Neural Networks for Virtual Axle Detection

\title{Object-Size-Driven Design of Convolutional Neural Networks: Virtual Axle Detection based on Raw Data}

%% use optional labels to link authors explicitly to addresses:
%% \author[label1,label2]{}
%% \affiliation[label1]{organization={},
%%             addressline={},
%%             city={},
%%             postcode={},
%%             state={},
%%             country={}}
%%
%% \affiliation[label2]{organization={},
%%             addressline={},
%%             city={},
%%             postcode={},
%%             state={},
%%             country={}}

\author[label0]{Henrik Riedel}\corref{cor1}
\author[label0]{Steven Robert Lorenzen}
\author[label0]{Clemens Hübler}

\cortext[cor1]{
    Corresponding author. \textit{E-mail address:} riedel@ismd.tu-darmstadt.de (H.Riedel).
    }
\affiliation[label0]{organization={Institute of Structural Mechanics and Design, Department of Civil and Environmental Engineering, Technical University of Darmstadt},%Department and Organization
            addressline={Franziska-Braun-Str. 3}, 
            city={Darmstadt},
            postcode={64287}, 
            state={Hesse},
            country={Germany}}

\begin{abstract} As infrastructure ages, the need for efficient monitoring methods becomes increasingly critical. Bridge Weigh-In-Motion (BWIM) systems are crucial for cost-effective determination of loads and, consequently, the residual service life of road and railway infrastructure. However, conventional BWIM systems require additional sensors for axle detection, which must be installed in potentially inaccessible locations or places that interfere with bridge operation.

This study presents a novel approach for real-time detection of train axles using sensors arbitrarily placed on bridges, providing an alternative to dedicated axle detectors. The developed Virtual Axle Detector with Enhanced Receptive Field (VADER) has been validated on a single-track railway bridge using only acceleration measurements, detecting 99.9~\% of axles with a spatial error of 3.69~cm. Using raw data as input outperformed the state-of-the-art spectrogram-based method in both speed and memory usage by 99~\%, thereby making real-time application feasible for the first time.

Additionally, we introduce the Maximum Receptive Field (MRF) rule, a novel approach to optimise hyperparameters of Convolutional Neural Networks (CNNs) based on the size of objects. In this context, the object size relates to the fundamental frequency of a bridge. The MRF rule effectively narrows the hyperparameter search space, overcoming the need for extensive hyperparameter tuning. Since the MRF rule can theoretically be applied to all unstructured data, it could have implications for a wide range of deep learning problems, from earthquake prediction to object recognition. \end{abstract}


\end{frontmatter}

%\begin{linenumbers}

\section{Introduction} \label{S:1} 

As infrastructure ages, efficient monitoring methods become increasingly imperative. Precise knowledge of the actual stress and loads on infrastructure enables more accurate determination of remaining service life, identification of overloaded vehicles, and more efficient planning for new constructions. Directly measuring loads is often challenging; instead, loads can be reconstructed via the structural response \citep{Chan2001,KazemiAmiri2017,Hwang2009,Lourens2012,Firus2021,Firus2022}. Weigh-In-Motion (WIM) systems are used to reconstruct the axle loads of trains or cars during regular traffic flow. However, conventional WIM systems must be installed directly into roadways or tracks, which often disrupts traffic, making them expensive and unsuitable for nationwide investigations of axle loads. In contrast, Bridge WIM systems (BWIM) are installed beneath bridge structures, allowing reuse and repositioning, which makes them cost-effective and suitable for time-limited investigations \citep{Lydon2017d,Wang2019,Yu2016b,He2019}.

\textbf{BWIM systems} are essential for large-scale investigations of axle loads but typically require knowledge of the axle positions for efficient use \citep{He2019,OBrien2012,Zhao2020}. For this purpose, additional sensors currently need to be installed at bridge-specific locations (facing the roadway, on the cross girders, or near the supports) \citep{Thater1998,Zakharenko2022,Bernas2018,Kouroussis2015}. The use of BWIM strain sensors for axle detection is limited to thin slab bridges \citep{Lydon2016}. \citet{JBENF2} proposed a BWIM system without axle detectors, but this system relies on strain measurements and only works for one vehicle at a time on the bridge. \citet{Kalhori2017} introduced a shear strain-based method requiring additional sensors with specific placement on the neutral axis of the bridge, while \citet{Zhu2021} developed a deep-learning-based method using accelerometers close to the bearings. For large-scale BWIM applications, a method for axle detection and localisation independent of bridge type, physical quantities, and sensor position would significantly reduce cost and risk. To the best of our knowledge, \citet{Lorenzen2022VirtualAD} is the only study to propose a method in which the sensors can be installed arbitrarily on the bridge without the need for additional sensors. To date, the feasibility of a virtual axle detector (VAD) has only been tested based on the assumption that spectrograms make it easier for the model to recognise the axles and that standard computer vision models do not require adaptation for spectrograms. Building on \citet{Lorenzen2022VirtualAD}, the innovations of this work include investigations regarding the representation of the input data, the model's performance on unseen train types, and how measurement data-specific engineering knowledge can enhance the model.

\textbf{Sound Event Detection} (SED) is defined in the literature as the detection of events in time series \citep{mesaros2021sound,8673582,9106322,9054708}. SED can, for example, be used for speech recognition. Here, the input is audible sound and events refer to words. This methodology can also be applied to other vibration signals like accelerations in bridges to detect events such as train axles passing a sensor. For SED problems, time series are typically transformed into 2D spectrograms and analysed using Convolutional Neural Networks (CNNs) \citep{Latif2020DeepRL,8678825} or, more recently, Transformers \citep{Radford2022RobustSR}, even though 1D CNNs have shown competitive performance \citep{KIRANYAZ2021107398}. Research often focuses on identifying the most effective spectrogram type \citep{Arias-Vergara2021,BOZKURT2018132}. In conventional tasks like speech recognition and acoustic scene classification, log mel spectrograms are predominant \citep{8673582,Mesaros2019AcousticSC,Radford2022RobustSR}, although studies show that using raw data can achieve comparable or superior results \citep{Oord2016WaveNetAG,Ghahremani2016AcousticMF,Sailor2016NovelUA}.

\citet{9686838} conclude that audio representation still heavily depends on the application. Spectrograms are praised for their visualisation capability and adaptability for sound perception, but their transformations often cannot be inverted directly. In contrast, raw data preserves all information but is said to require more computing power \citep{9686838}. The higher demand for computing power stems from the fact that raw signals are often downsampled before being transformed into spectrograms \citep{Radford2022RobustSR}. However, with the same resolution in the time domain, spectrograms are inherently larger and thus more computationally intensive than raw signals. The good visualisation capability of spectrograms simplify feature extraction since they are interpretable by humans, enabling the definition of manual rules (e.g., peak finding in a certain frequency range). Conventional machine learning models, such as support vector machines or decision trees, cannot process unstructured data directly \citep{geron}, making feature extraction a prerequisite for their application. This is not a prerequisite anymore for new deep learning methods. Here, spectrograms likely remained the preferred type of input data because they allowed the use of image-based networks without the need for modifications. However, it remains unclear whether spectrograms still provide an advantage for deep learning models that have been optimised for raw signals.

In the context of axle localisation, Continuous Wavelet Transformations (CWTs) have proven effective as features for axle detection \citep{Chatterjee2006,Kalhori2017,Yu2017,Zhao2020,Zhu2021,Lorenzen2022VirtualAD}. Given their strong performance in speech recognition \citep{Arias-Vergara2021} and the tailoring of log mel spectrograms to human auditory perception, \citet{Lorenzen2022VirtualAD} previously adopted CWTs for axle detection. While \citet{Lorenzen2022VirtualAD} demonstrated the feasibility of a virtual axle detector using a CNN model, the study did not explore the impact of model design, data representation (raw or spectrogram), and receptive field (RF) size. In addition, it remains unclear whether converting data into 2D spectrograms achieves superior results. To our knowledge, no study has systematically optimised models for both raw data and spectrograms to directly compare their outcomes.

\textbf{Hyperparameter tuning} and model architecture optimisation are typically based on empirical values or search algorithms, with few established guidelines \citep{geron}. While strategies exist for selecting the learning rate, the number of epochs, and the mini-batch size, no clear approach is available for determining the number of convolutional layers or the size and number of convolutional kernels \citep{CNNoptimization,CNNsurvey}. \citet{VGG} showed that large kernels are unnecessary, as successive convolutional layers with smaller kernels can achieve the same effective receptive field (ERF). Further studies have investigated ERF size with images \citep{understandingERF,understandingERFIsegmentation} and with audio signals \citep{RFregularizationAudio}, finding a correlation between ERF size and model performance. However, it remains unclear why specific ERF sizes are better and how to leverage them for model development.

Until the necessary ERF size for a given problem is known, hyperparameter tuning remains essential. \citet{CNNarchis} suggest that hyperparameter selection requires intuition or can be performed by genetic algorithms. Tools like Keras, Optuna, and Hyperopt \citep{chollet2015keras,optuna,Hyperopt}, along with research on neural architecture search \citep{neuralArchiSearch1000,neuralArchiSearch,neuralArchiSearchHardware}, employ reinforcement learning or evolution to empirically optimise hyperparameters \citep{CNNcomprehensiveSurvey}. However, this approach results in additional computing and energy costs. A derivable rule with a theoretical basis could significantly narrow the search space, saving time and energy. In this study, we propose the novel Maximum Receptive Field (MRF) rule, based on object sizes, and validate it using the virtual axle detector. Here, the object size refers to the minimum fundamental frequency of a bridge, but the basic principle is transferable to other domains such as sound or images. For instance, in semantic image segmentation, the object size could correspond to the largest image section required to identify an object \citep{crack}. In earthquake monitoring, it might refer to the frequencies above 1Hz, which are important for monitoring micro-earthquakes \citep{earthquake}, and in speech to frequencies in the range of 80--300Hz, which dominate human speech \citep{speech}. The sampling rate would then be used to convert the frequencies to the number of samples needed to optimise the ERF. This approach has broad potential and can be applied wherever the object sizes can be derived by humans using domain knowledge.

\textbf{In this study}, we propose a novel approach addressing the research gap of real-time axle detection and localisation using raw data of arbitrarily placed sensors. For each sensor, the axles are localised at the nearest point of the tracks to the sensor (perpendicular to the direction of travel). When using multiple sensors, multi-dimensional positioning is possible, extending the concept to car bridges as well. This enables BWIM systems to conduct real-time axle detection without additional sensors. The concept is validated on a single-track railway bridge, with one train at a time on the ballasted track. To make it transferable to car bridges, the axles were always analysed individually rather than as part of a vehicle.

The proposed concept determines the longitudinal position of axles using sensors distributed longitudinally along the bridge. Additional transversally distributed sensors could provide two-dimensional localisation, making the approach applicable for multi-lane car bridges. To validate our approach, we adapted the VAD model \citep{Lorenzen2022VirtualAD} to handle raw measurement data and tested it on the dataset from \citet{lorenzen_steven_robert_2022_6782319}. Additionally, we investigated the best input data representation, performance of the model on unseen train types, and model optimisation through data-specific engineering knowledge. Finally, we propose a novel MRF rule based on object sizes for optimising CNN models with time-series, image, or video data for various applications.
\section{Methodology}\label{S:2}

In this section, we outline the methodology used for the Virtual Axle Detector with Enhanced Receptive Field (VADER) and compare how data representation impacts it. A summary of the dataset \citep{lorenzen_steven_robert_2022_6782319} is provided, followed by details on the training parameters and evaluation metrics.

\subsection{Data Acquisition}
\begin{figure*}
\centering\includegraphics[trim = 0mm 40mm 185mm 13mm,clip,width=\textwidth]{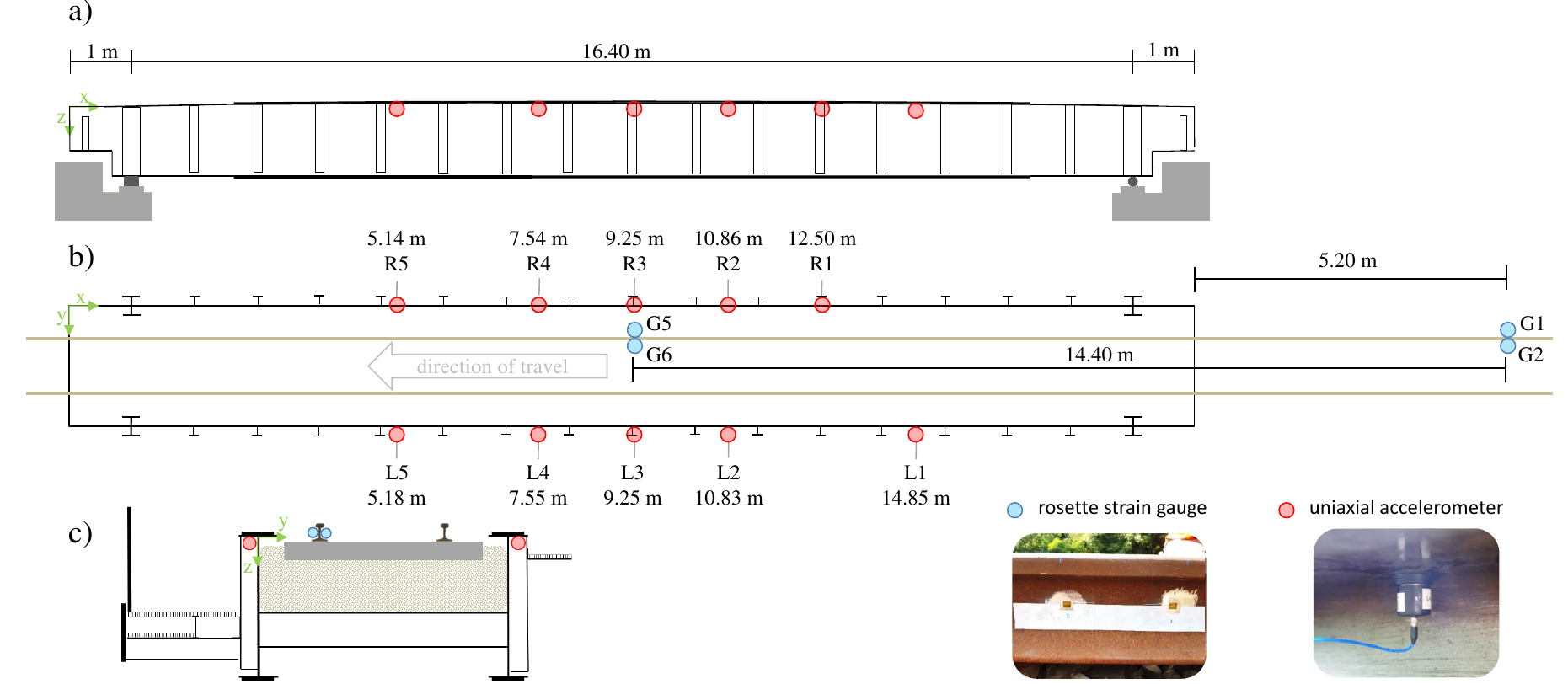}
\caption{Bridge and sensor setup: (a) side view; (b) top view with sensor labels, accelerometers $x$-ordinate and strain gauge distances; (c) cross section. \citep{Lorenzen2022VirtualAD}}
\label{F:BridgeSensorsetup}
\end{figure*}

The dataset from \citet{lorenzen_steven_robert_2022_6782319} was collected on a single-span steel trough railway bridge located on a long-distance traffic line in Germany (Fig.~\ref{F:BridgeSensorsetup}). This bridge is 18.4 meters long with a free span of 16.4 meters and a fundamental frequency of about 6.9~Hz for the first bending mode. The fundamental frequency varies non-linearly with the load due to the ballast \citep{reiterer2022dynamische}. Data was collected from ten seismic uniaxial accelerometers installed along the bridge using a sampling rate of 600~Hz.

\begin{figure*}
    \centering
    \includegraphics[width=0.75\textwidth]{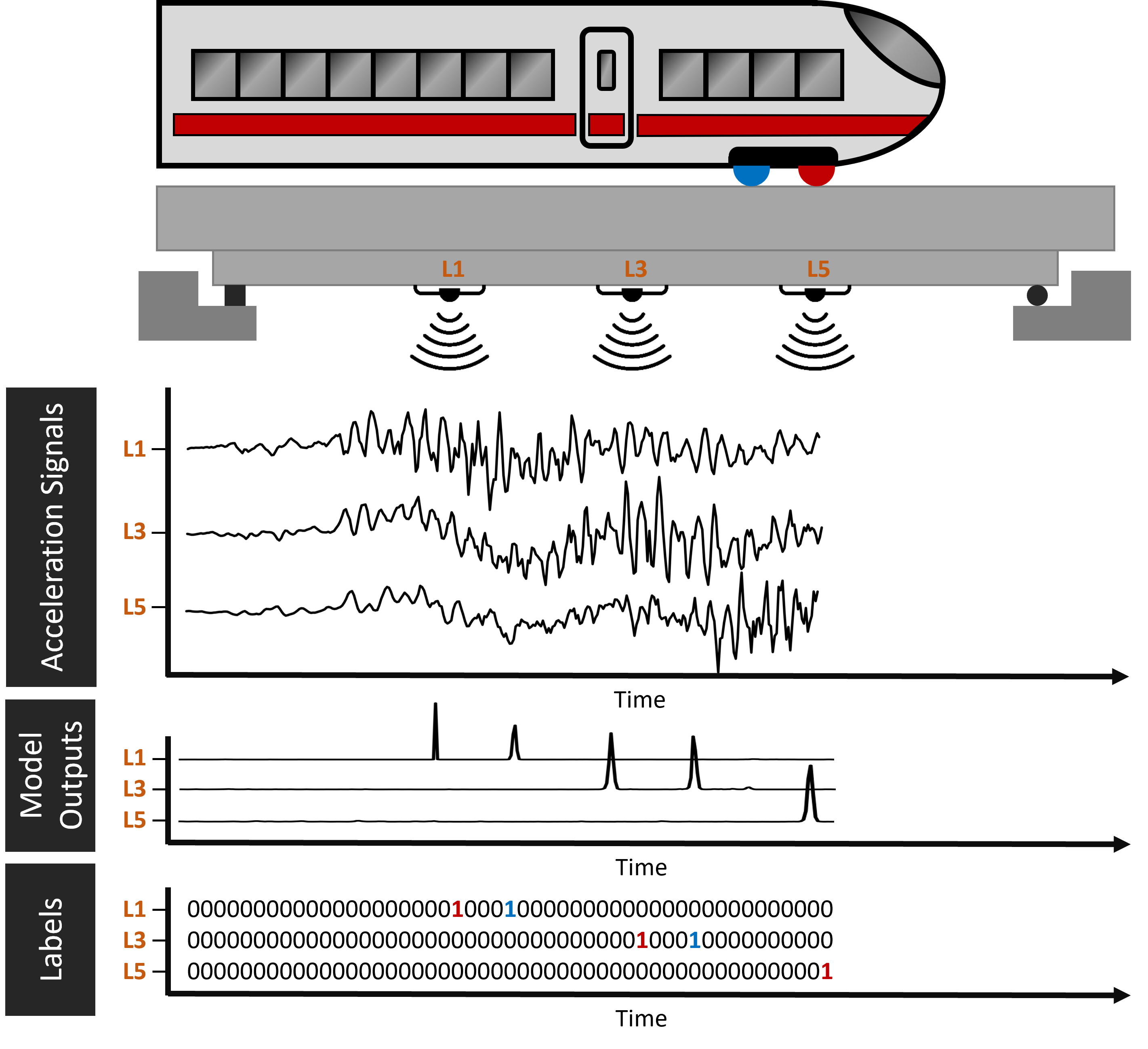}
    \caption{Measured acceleration signals of sensors L1, L3, and L5 (Fig.~\ref{F:BridgeSensorsetup}) with corresponding outputs of a VADER model. The labels represent the time point when a train axle is above the respective sensor. The axle and the corresponding label are shown in red for the first and in blue for the second axle. Measured acceleration signals are shown with the corresponding model outputs. Labels, bridge, sensors, and train are shown in simplified versions. The models generally output probabilities between zero and one. If the model is uncertain, a high probability for a passing axle is predicted over several time steps. To compare the model predictions with the labels, peak picking was used to select the time point when the model is most certain. A peak is only picked if the model's confidence is at least 25~\% and there is a minimum of 20 samples ($0.0\overline{3}$~seconds) between consecutive peaks.}
    \label{fig:label}
\end{figure*}

Using a supervised learning approach, our model requires ground truth labels for training. The labels were generated using four rosette strain gauges directly installed on the rails (Fig.~\ref{F:BridgeSensorsetup}), ensuring sufficient accuracy to evaluate the model's performance. The axle positions and velocities of the passing trains were derived from the rosette strain gauges and were used to determine the times when the axles were closest to each acceleration sensor (perpendicular to the direction of travel). The labels are then created using these calculated points in time. The label vector contains ones at the times when a train axle is above the respective sensor and zeros otherwise (Fig.~\ref{fig:label}). As a result, the model is trained to use the acceleration sensors as a virtual light barrier, triggered whenever a train axle passes \citep{Lorenzen2022VirtualAD}. The sum of each label vector corresponds to the number of axles of the train. The model only receives acceleration signals from a single sensor at a time and predicts the times at which an axle is located above the sensor (Fig.~\ref{fig:label}). Since the signals from the sensors are similar, especially the low-frequency component, the model must also infer which sensor (position) recorded the data. Using two or more acceleration sensors, the axle distances and velocities can be determined. Due to the 600~Hz sampling rate, label inaccuracies ranging from 6~cm to 38~cm were assumed, depending on train velocity and the sensor position. The dataset comprises 3,745 train passages, with one label vector per sensor. For more details on the dataset and the measurement campaign, please see \citet{Lorenzen2022VirtualAD}.

To eliminate the need for strain gauge measurements, this study investigates two scenarios for labelling: (a) utilising data from a failed axle detector and (b) using trains with a differential global positioning system (DGPS).

\subsection{Data Transformation}
To investigate the impact of data representation, two approaches are considered: (a) spectrograms and (b) raw measurement data. The spectrograms are identical to those proposed by \citet{Lorenzen2022VirtualAD}, with three mother wavelets and two sets of 16 frequencies per mother wavelet as input per crossing and sensor (Table~\ref{tab:cwtsettings}). This results in an input tensor of $16 \times 6 \times n_\mathrm{s}$, with $n_\mathrm{s}$ being the number of samples in the acceleration signal of a sensor. In contrast, the raw data was used as input without any processing, scaling, or normalisation.

\begin{table}
\begin{center}
\begin{tabular}{ l | c c } 
    & \multicolumn{2}{c}{Scale Limit} \\
    Wavelet & Lower & Upper \\
    \hline \hline
    \multirow{2}{12em}{First Order Complex Gaussian Derivative} & 1 & 8 \\ 
    & 8 & 50 \\ 
    \hline
    \multirow{2}{12em}{First Order Gaussian Derivative} & 0.6 & 6.5 \\ 
    & 6.5 & 35 \\ 
    \hline
    \multirow{2}{12em}{Default Frequency B-Spline from \citet{Lee2019}} & 1.5 & 10 \\ 
    & 10 & 40 \\ 
    
\end{tabular}
\end{center}
\caption{The transformation settings used per passage and per signal, derived from \citet{Lorenzen2022VirtualAD}.}
\label{tab:cwtsettings}
\end{table}

\subsection{Data Split}

To evaluate the performance and generalisation capability of the proposed VADER model, a five-fold cross-validation approach with a holdout test set was employed. Initially, the dataset was divided into two main parts: a test set and a remaining set for training and validation. The test set was held out and used exclusively for the final evaluation of the model, ensuring an unbiased assessment on unseen data.

The remaining data was then partitioned into five folds of equal size. In each iteration of the cross-validation process, one fold served as the validation set, while the other four folds were combined to form the training set. Training was stopped based on the performance on the validation set, and after each training phase, the model was evaluated on the same held-out test set. This process allowed the model to be evaluated five times using different training and validation sets, providing a more accurate estimation of its generalisation capability \citep{geron}.

\begin{figure*}
    \centering

\begin{subfigure}{0.49\textwidth}
    \centering
    \includegraphics[width=\linewidth]{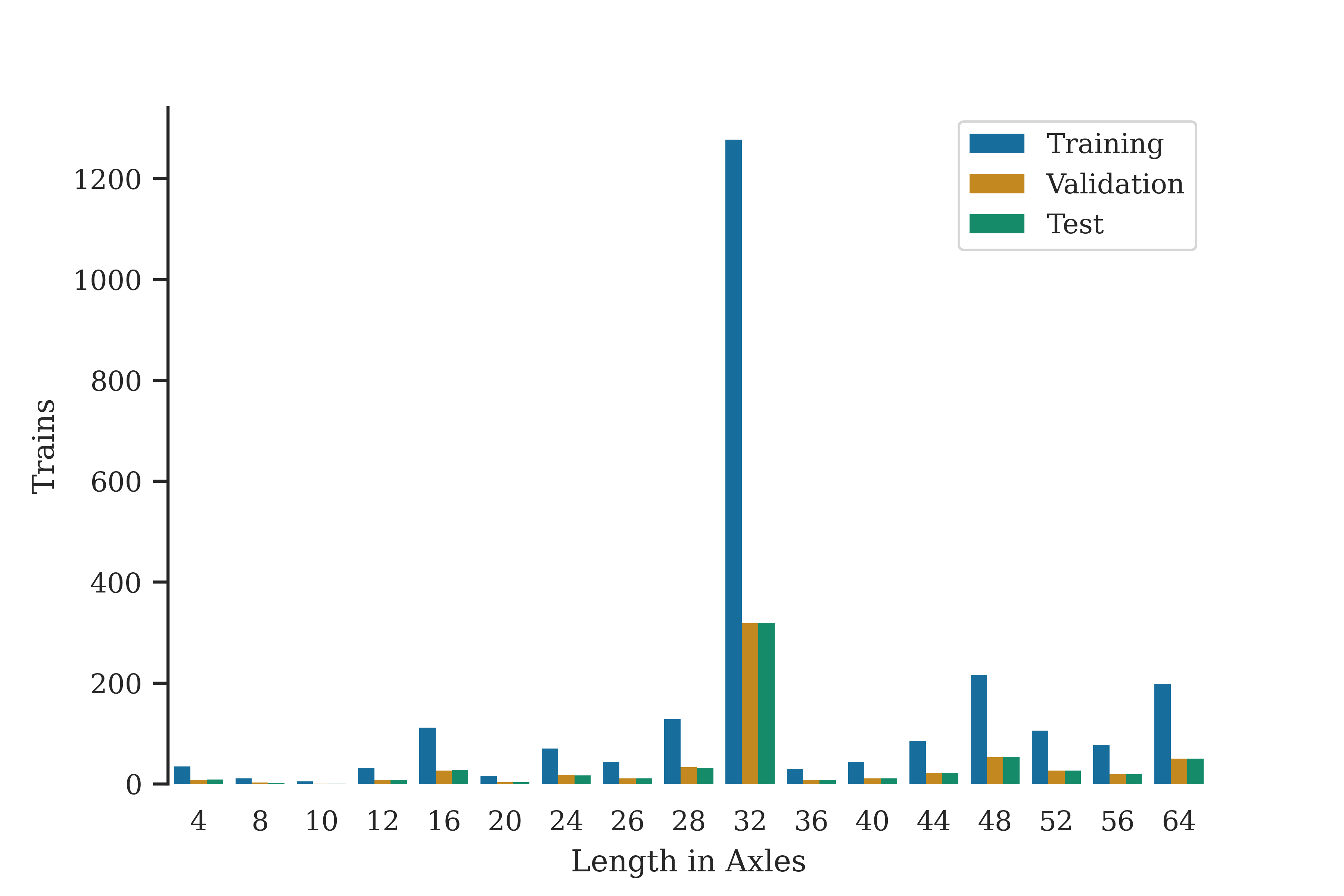}
    \caption{First fold of the stratified scenario}
    \label{fig:dataset_stratified}
\end{subfigure}
\hfill
\begin{subfigure}{0.49\textwidth}
    \centering
    \includegraphics[width=\linewidth]{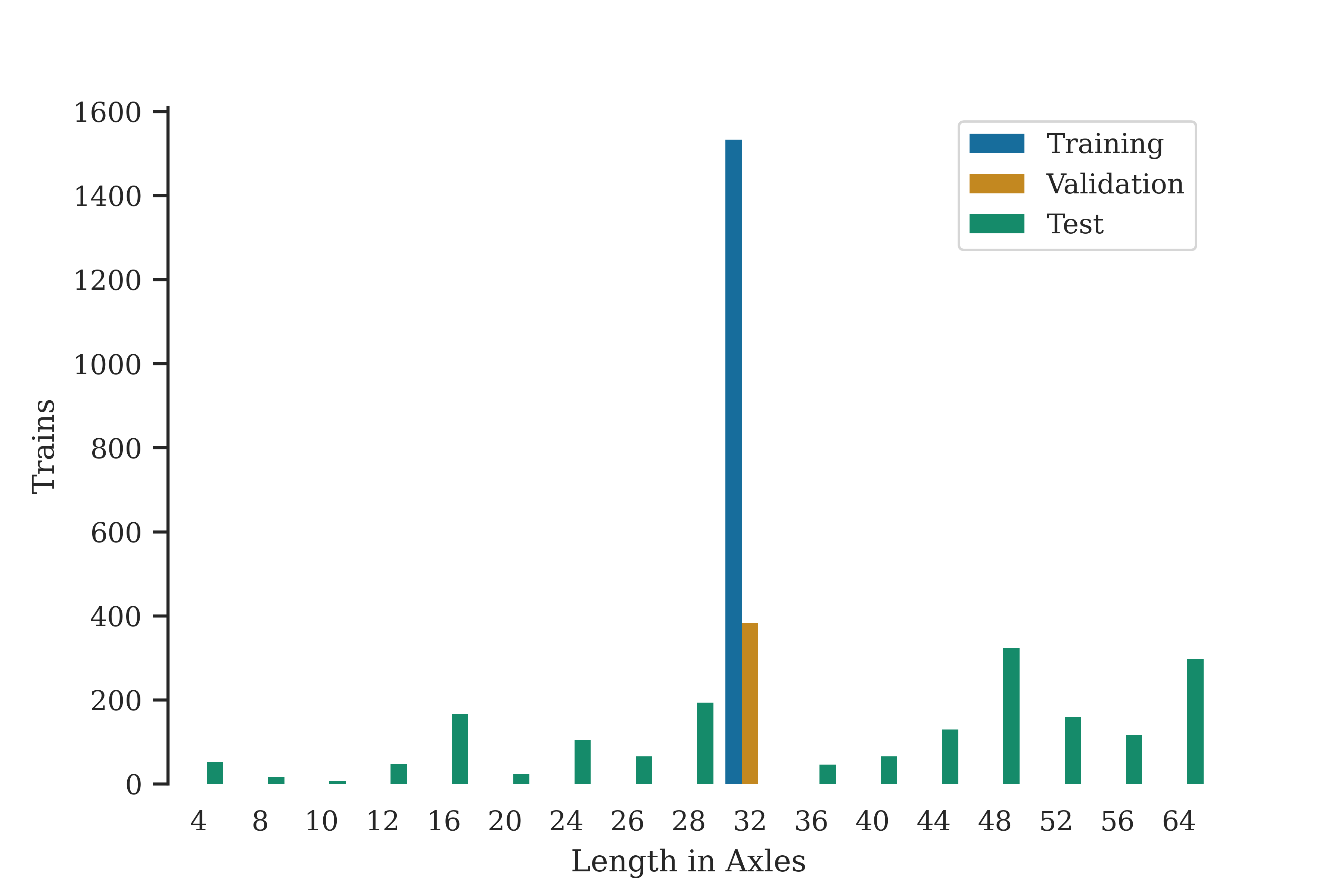}
    \caption{First fold of the DGPS scenario}
    \label{fig:dataset_single}
\end{subfigure}

    \caption{Histogram of train lengths defined by number of axles for the first fold of both scenarios using five-fold cross-validation.}
    \label{fig:splits}
\end{figure*}

Since the exact train types are unknown, the train length in axles was used as a proxy for train type classification. To further investigate the applicability and robustness of the proposed concept, two different scenarios were considered:

\begin{enumerate}
    \item \textbf{Stratified Scenario:}
    This scenario simulates the situation where an existing axle detector has failed and needs to be replaced by VADER. We assume that we have correctly labelled data from when the axle detector was functioning properly. To reflect this, we performed a stratified split of the dataset based on train lengths (number of axles), ensuring that both the training and test sets accurately represent the overall distribution of train types. The split resulted in 3,110 passages with 112,038 individual axles for training and validation, and 623 passages with 22,472 individual axles for testing in each fold. This approach allowed us to assess how well VADER can replace an existing axle detector using data representative of normal operating conditions.

    \item \textbf{DGPS Scenario:}
    In the second scenario, we mimic a situation where only one type of train is equipped with a differential global positioning system (DGPS), providing accurately labelled data only for that train type.  Trains with the most common axle count were selected to form the training set. The remaining trains with different axle counts were included in the test set (Fig.~\ref{fig:dataset_single}). This results in 1,916 passages with 61,312 individual axles for training and 1,817 passages with 73,198 individual axles for testing. In this scenario, less data was available for training compared to the stratified scenario. This approach tested the model's ability to generalise from a non-representative training set to a diverse test set, highlighting its robustness under challenging conditions. 
\end{enumerate}

By using five-fold cross-validation in both scenarios, we ensured that the evaluation of VADER was thorough and that the results were not dependent on a particular data split. This methodology provided a comprehensive understanding of the models performance in both ideal conditions (stratified scenario) and challenging conditions (DGPS scenario).

\subsection{Model Definition}

The VADER model is implemented as a fully convolutional neural network (FCN) \citep{Long2015FullyCN} to handle inputs of arbitrary lengths. This property is particularly advantageous for the detection of train axles, as the crossing time varies significantly based on the train's speed and the number of axles.

The U-Net concept \citep{ronneberger2015unet} was chosen for the model architecture. It was originally developed for semantic segmentation of images but is also adaptable for segmenting time-series data. Despite the development of new architectures, the U-Net architecture remains the most widely used model for semantic segmentation \citep{unet_pwc} and continues to be applied across a variety of domains \citep{MO2022626,electronics12051199,9446143,freeUnet,medicalUnet,unifiedUnet}. Since U-Net serves as the foundation for many other models, it is also one of the simplest architectures. Consequently, the influence of the architecture on the results should be minimised, allowing for more general insights into the functionality of CNNs. Therefore, in this study, the effects of the hyperparameters kernel size, max pooling size, and pooling steps are tested on the U-Net. Investigations on other models are beyond the scope of this study.

\begin{figure*}
    \centering

\begin{subfigure}{\textwidth}    
\centering
    \includegraphics[width=\linewidth, trim= 46 68 8 0,clip]{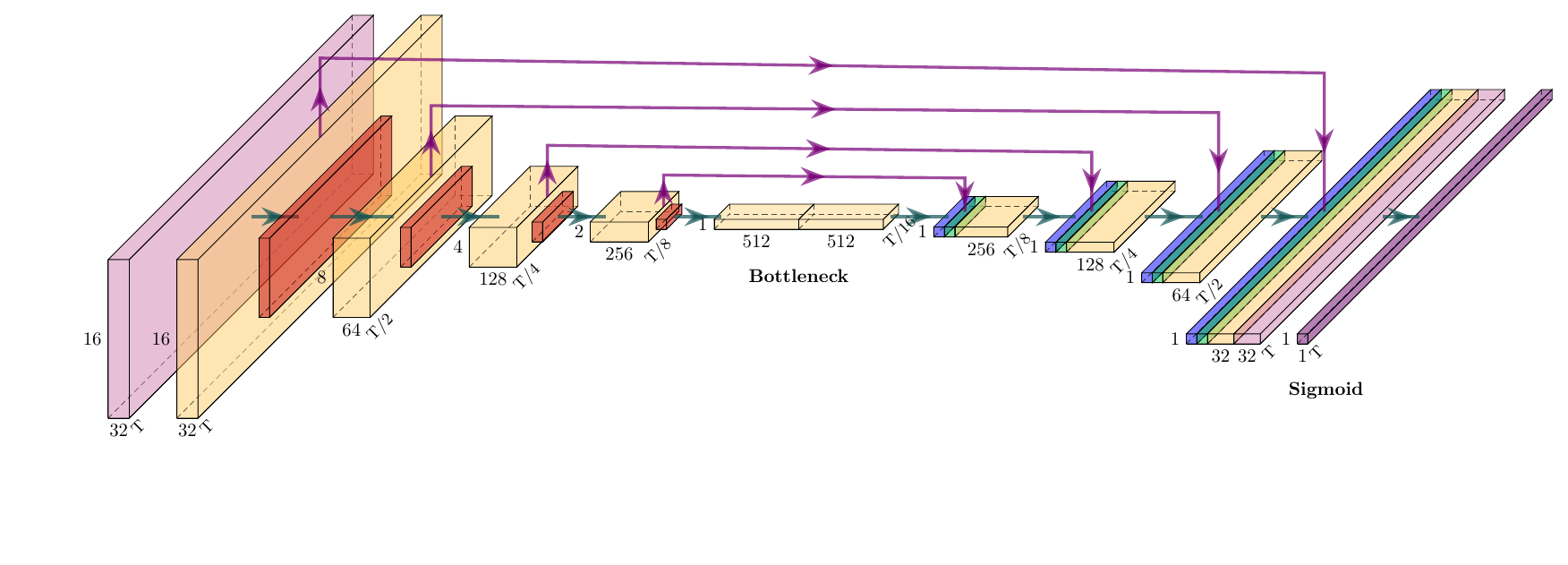}
    \caption{VADER with spectograms (\(\text{VADER}_\text{spec}\)) as input, max pooling size of 2 and 4 pooling steps. Derived from \citet{Lorenzen2022VirtualAD}.}
    \label{fig:VAD_model}
\end{subfigure}

\begin{subfigure}{\textwidth}  
    \centering
    \includegraphics[width=\linewidth, trim= 46 68 8 0,clip]{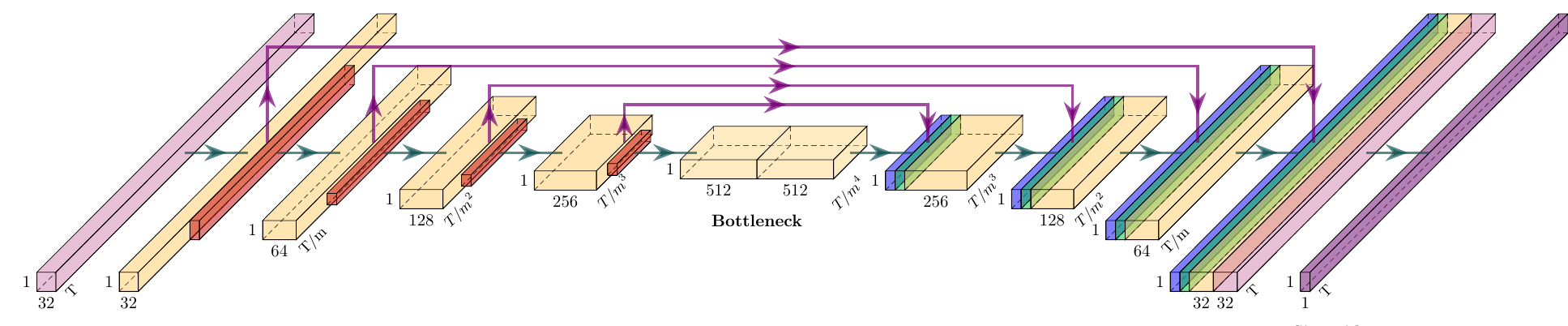}
    \caption{VADER with raw data (\(\text{VADER}_\text{raw}\)) as input and 4 pooling steps.}
    \label{fig:VADER_model}
\end{subfigure}

\begin{subfigure}{\textwidth}  
    \centering
    \includegraphics[width=0.87\linewidth, trim= 46 68 8 0,clip]{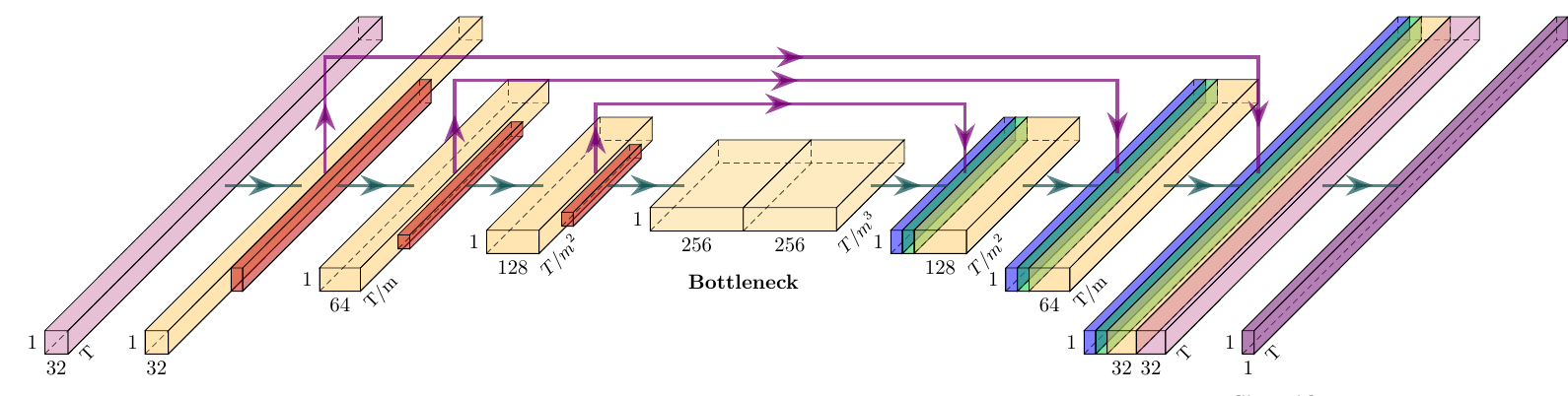}
    \caption{VADER with raw data (\(\text{VADER}_\text{raw}\)) as input and 3 pooling steps.}
    \label{fig:VADER_model_3}
\end{subfigure}

    \caption{Architecture of the U-Net-based VADER model. The network follows a symmetric encoder-decoder structure, where the left side represents the contracting path (encoder) with multiple convolutional layers (CB in light purple or RB in yellow) followed by max pooling (red) for down-sampling. The right side depicts the expansive path (decoder) with up-sampling operations (transposed convolution in blue), convolutional layers, and concatenation (green) with corresponding layers from the encoder via skip connections (purple arrow). The final layer applies a convolution with sigmoid activation (purple) to generate the output. The skip connections ensure high-resolution features from the encoder are preserved and integrated into the decoder. The boxes qualitatively represent the size of the outputs from the corresponding layer (feature maps) with \(T\) for samples and \(m\) for max pooling size at the bottom right, feature maps at the bottom and frequencies at the left. Three of the hyperparameter combinations examined were presented as examples.}
    \label{fig:models}
\end{figure*}

The fundamental concept of the U-Net architecture is the encoder and decoder paths. In the encoder path, the resolution is progressively reduced and information is compressed, while in the decoder path, the resolution is increased back to the original dimensions (Fig.~\ref{fig:VAD_model}). The encoder path aims to capture details at various resolutions, while the decoder path integrates these details into the appropriate context. The intermediate results from the encoder and decoder paths at the same resolution are combined through skip connections.

The pooling layer in U-Net plays a key role in progressively reducing the resolution of feature maps, thereby enabling the model to capture more abstract features and increasing the MRF. This process of downsampling helps to focus the network on the most important features, even though some fine details might be lost. However, these details are later recovered during the upsampling process in the decoder path, aided by the skip connections that transfer the corresponding high-resolution features from the encoder path. Strided convolutions are often used instead of max pooling as a subsampling method to decrease computational cost. However, it was shown that max pooling achieves one of the highest accuracies compared to other subsampling methods \citep{poolingBenchmark}. It also has a slightly higher accuracy than strided convolutions but with a longer training and inference time \citep{poolingVsStride}. For these reasons, and because the concept of max pooling is one of the more easily understood subsampling concepts, only max pooling is used as the subsampling method in this study.

Our implementation of the U-Net comprises a well-established combination \citep{geron} of convolution blocks (CBs), residual blocks (RBs) originally proposed by \citet{he2015deep}, max pooling layers, concatenate layers, and transposed convolution layers (Fig.~\ref{fig:models}). A CB always consists of a convolution layer with Rectified Linear Unit (ReLU) activation and a group normalisation layer \citep{Wu2018GroupN}. The first group normalisation layer has a single group (making it nearly identical to Layer Normalisation \citep{Wu2018GroupN}), and the remaining group norm layers have 16 channels per group. As a result, the measurement data is first normalised individually (as 1 channel per sensor) in the first group normalisation layer, and the remaining CNN feature maps are normalised with the optimum group size according to \citet{Wu2018GroupN}. Group normalisation was shown to be less sensitive to other hyperparameters like batch size and was therefore chosen to effectively reduce the hyperparameter space. The RBs were implemented similarly to the 50-layer or larger variants of the ResNet \citep{he2015deep}. The final layer in both models is a convolution layer with only one filter and sigmoid activation.

The VADER model has been implemented parametrically so that it can process either spectrograms (\(\text{VADER}_\text{spec}\)) or raw data (\(\text{VADER}_\text{raw}\)) as input. In addition, three hyperparameters can be varied: the number of pooling steps, the pooling size, and the kernel size. In each case, the kernel size is at most as large as the corresponding dimension. For a selected kernel size of nine, the kernel for raw data is therefore always $9 \times 1$. For spectrograms, with a selected kernel size of nine, the first layer is $9 \times 9$ and the second layer is $9 \times 8$ for a pooling size of two, because the frequency range has been reduced from 16 values to eight values.

The MRF refers to the number of samples that the kernel can see at the original resolution or sampling rate and is defined in this study as follows:

\begin{equation}\label{eq:MRF}
    \text{MRF} = k \times m^{p}
\end{equation}

with $k$ as kernel size, $m$ as max pooling size, and $p$ as pooling steps. After a max pooling layer of size two, the MRF would therefore double with the same kernel size. In order to determine the optimal pooling size, kernel size, and amount of pooling steps, we propose the MRF rule: For a CNN to achieve good results, its MRF must be at least as large as the largest object of interest. In the case of acceleration signals, the object size $y$ can be determined via the sampling frequency $f_\mathrm{s}$ (in Hz) and the lowest frequency of interest $f_\mathrm{l}$ (in Hz):

\begin{equation}\label{eq:objectsize}
    y = \frac{f_\mathrm{s}}{f_\mathrm{l}}
\end{equation}

This means that a signal with a frequency of, e.g., 1~Hz at a sampling rate of 600~Hz has an object size of 600~samples. When inserted into the MRF rule, this results in a system of equations with three variables (kernel size, max pooling size, and pooling steps):

\begin{align}
    y &\leq \text{MRF}, \label{eq:MRFvsObjectsize}\\
    \frac{f_\mathrm{s}}{f_\mathrm{l}} &\leq  k \times m^{p} \label{eq:final}
\end{align}

The bridge has a fundamental frequency of up to 6.9~Hz (depending on the amplitude), and to distinguish the contribution of the bridge from the axle loads' structural response, an MRF of at least $\frac{600~\mathrm{Hz}}{6.9~\mathrm{Hz}} \approx 87$~samples would be necessary. Taking into account that the resonance frequency decreases to about 5.6~Hz for big amplitudes \citep{LorenzenDiss}, an MRF of $\frac{600~\mathrm{Hz}}{5~\mathrm{Hz}} = 120$~samples ensures that more than the largest object size is captured.

\begin{figure*}
    \centering
    
    \begin{subfigure}{0.49\textwidth}
    \centering
    \includegraphics[width=\linewidth]{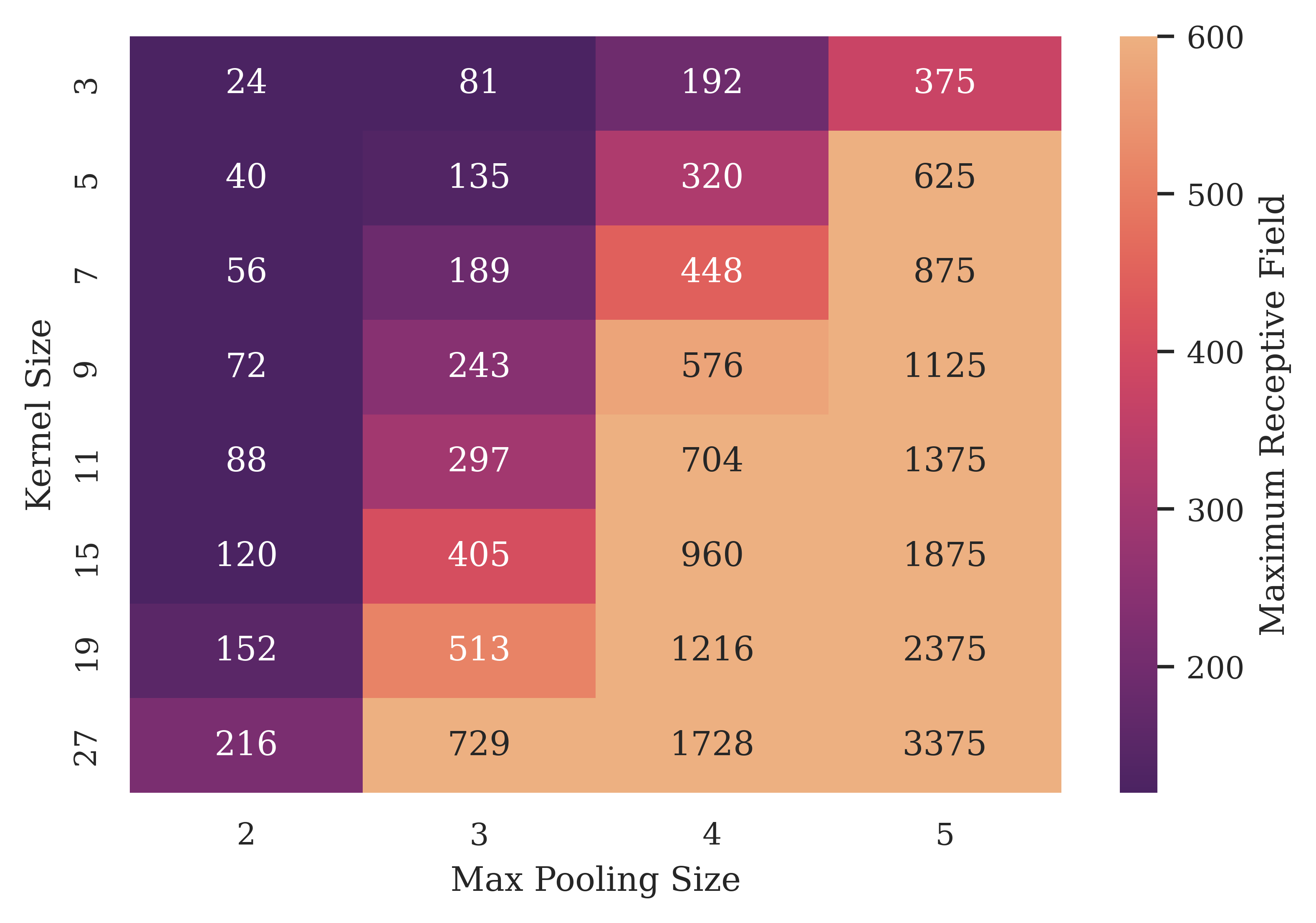}
    \caption{Resulting maximum receptive fields for \(\text{VADER}_\text{raw}\) with 3 pooling steps.}
    \label{fig:RF_MP_3}
    \end{subfigure}
\hfill
    \begin{subfigure}{0.49\textwidth}
    \centering
    \includegraphics[width=\linewidth]{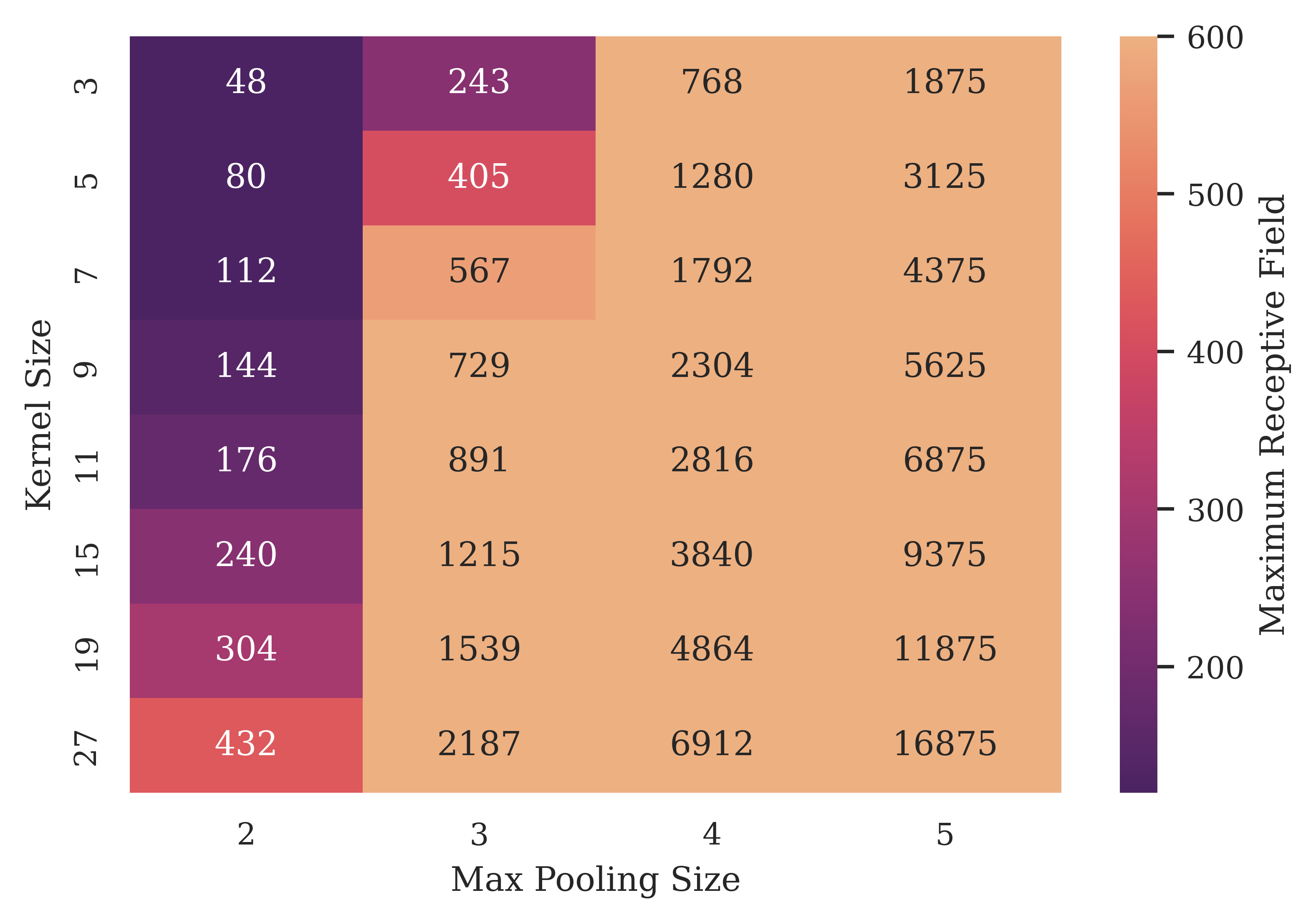}
    \caption{Resulting maximum receptive fields for \(\text{VADER}_\text{raw}\) with 4 pooling steps.}
    \label{fig:RF_MP_4}
    \end{subfigure}
    
    \caption{Maximum receptive fields as a function of kernel size, max pooling size, and pooling steps for all used combinations of the hyperparameter study.}
    \label{fig:parameter_study}
\end{figure*}

Much like CNNs, wavelets convolve over the samples, effectively encoding frequency information into individual sample values through the CWT transformation. Depending on the wavelet and the scale, different frequencies are filtered. With the parameters used, information on frequencies up to approximately 3.4~Hz is therefore encoded in individual samples. Therefore, the CWT transformation can effectively increase the MRF for \(\text{VADER}_\text{spec}\) to $\frac{600~\mathrm{Hz}}{3.4~\mathrm{Hz}} \approx 176$~samples. However, since the RF is effectively only increased for the low-frequency part, it is unclear what effect it has on the overall performance.
To investigate the effect of the MRF on the results, two amounts of pooling steps, eight different kernel sizes, and four different max pooling sizes were examined for both data representations as a hyperparameter study (Fig.~\ref{fig:parameter_study}).

Depending on the hyperparameter combinations, there are large differences in the maximum receptive field size of the respective model (Fig.~\ref{fig:parameter_study}). With a required MRF size of about 87--120~samples (depending on whether the fundamental frequency is considered constant or not), a small proportion of the models are expected to perform significantly worse (dark purple in Fig.~\ref{fig:parameter_study}). From an MRF size above $\frac{600~\mathrm{Hz}}{1~\mathrm{Hz}} \approx 600$~samples, there is expected to be no further improvement in the results (light orange in Fig.~\ref{fig:parameter_study}). There are no objects with frequencies of less than 1~Hz that should be able to help the model recognise train axles. An MRF that can capture frequencies of 1~Hz already allows capturing several bridge oscillation periods in a single RF. Beyond that, the risk increases that entire trains and thus axle configurations are learned by the model. This could therefore increase the probability of overfitting and reduce the generalisation capability.

The TensorFlow library \citep{tensorflow2015-whitepaper} was used for implementation, and PlotNeuralNet \citep{haris_iqbal_2018_2526396} for visualisation of the models.

\subsection{Training Parameters}

For the loss function, binary focal loss was chosen with an optimised \(\gamma\) value of 2.5 \citep{Lorenzen2022VirtualAD}, as it is particularly suitable for imbalanced datasets \citep{Lin2017FocalLF}. Adam, originally proposed by \citet{Kingma2014AdamAM}, was used as the optimiser with the default initial learning rate of 0.001. To evaluate the model's performance, the \(F_1\) score (Eq.~\ref{eq_f1}) was used as the standard metric for imbalanced datasets \citep{geron}:

\begin{equation}  \label{eq_f1}
    F_1 = 2 \times \frac{\text{precision} \times \text{recall}}{\text{precision} + \text{recall}} \times 100
\end{equation}

where

\begin{align}    
    \text{precision} &= \frac{\text{TP}}{\text{TP} + \text{FP}} \\
    \text{recall} &= \frac{\text{TP}}{\text{TP} + \text{FN}}
\end{align}

with TP as the number of true positive results, FP as the number of false positive results, and FN as the number of false negative results.

To determine when an axle is considered correctly detected (TP), two spatial threshold values were defined. The first threshold of 200~cm corresponds to the minimum assumed distance between two axles, and the second threshold of 37~cm corresponds to the maximum expected error in label creation. Training was conducted for an arbitrary maximum of 300 epochs with a batch size of 16. Within an epoch, all training data is iterated through once. After three epochs without improvement in the \(F_1\) score on the validation set, the learning rate was reduced by a factor of 0.3 (the default factor of 0.1 slowed down training too much), and the training was terminated after six epochs without improvement. For the hyperparameter study, each model was trained once instead of five times on the folds and the test data of the stratified splits.

The second metric to be determined is the mean absolute spatial error \(\overline{\Delta s}\) between prediction and labelling. For this purpose, the temporal error is first determined in the form of the number of samples between the prediction \(\hat{t}\) and the label \(t\), and this is then multiplied by the respective axle velocities:

\begin{equation}  \label{eq:SE}
    \overline{\Delta s} = \frac{1}{n} \sum_{i=1}^{n} \lvert t_i - \hat{t}_i \rvert \times v_i
\end{equation}

where \(n\) is the number of axles, \(v_i\) is the velocity in meters per sample, \(t_i\) is the predicted crossing time in samples, and \(\hat{t}_i\) is the ground truth crossing time in samples for axle \(i\).

The \(\overline{\Delta s}\) is converted into mean spatial accuracy (MSA). The minimum spatial accuracy was set at two meters (as one of the thresholds for the \(F_1\) score). This means that only detected axles with a \(\overline{\Delta s}\) of less than two meters are taken into account for MSA. As a result, the second metric is transformed into the same scale as the \(F_1\) score and can be compared more intuitively:

\begin{equation}  \label{eq:SA}
\text{MSA} = \frac{\text{ST} - \overline{\Delta s}}{2} \times 100
\end{equation}

with ST as the spatial threshold of 200~cm. The last metric used is the harmonic mean (HM) of the \(F_1\) score and the MSA, to evaluate both metrics together. While one metric could dominate the overall result with the arithmetic mean, both metrics must be high to achieve a high HM value in both scenarios. For two metrics with values of 99~\% and 1~\%, this results in an arithmetic mean of 50~\% and an HM of 1.98~\%. This ensures that the selected model performs well in both metrics and both scenarios. HM is defined here as follows:

\begin{equation} \label{eq:hm}
\text{HM} = \frac{4}{\frac{1}{\text{MSA}_{\text{stratified}}} + \frac{1}{\text{MSA}_{\text{DGPS}}} + \frac{1}{F_{1,\text{stratified}}} + \frac{1}{F_{1,\text{DGPS}}}}
\end{equation}

\section{Results and Discussion}
\label{S:3}

In this section, we present the results of our study on real-time train axle detection using bridge-mounted sensors. We evaluate the computational efficiency of our proposed method, \(\text{VADER}_\text{raw}\), compared to spectrogram-based approaches, demonstrating significant improvements. We discuss the findings from the hyperparameter study, highlighting the impact of the MRF size on model performance. Validation and test results are analysed to assess the generalisation capabilities of the models across different scenarios and sensor configurations. Finally, we examine the sensor-dependent results and confirm the effectiveness of using raw acceleration data and the proposed MRF rule for optimising convolutional neural networks in this application.

\subsection{Computational Efficiency}

Computational efficiency is critical in the context of axle detection for railway bridges, especially for real-time monitoring systems where trains pass frequently. Using raw data instead of Continuous Wavelet Transforms (CWTs) significantly reduces computation time and memory requirements, making real-time processing feasible.

For the six CWTs with 16 scales each, as proposed in \citet{Lorenzen2022VirtualAD}, processing a 12-second measurement requires 1.4~seconds and about 5.7~MB of memory just for the CWT computation. In contrast, our approach using raw data takes only 0.022~seconds from raw data input to prediction, with about 61.1~KB of memory usage. Therefore, when using raw data, inference becomes at least 65 times faster compared to using multi-channel spectrograms, while utilizing only about 1~\% of the memory for the input.

To illustrate this difference in a practical scenario, consider axle detection for a train with 64 axles using 10 acceleration sensors. Using \(\text{VADER}_\text{spec}\), the processing time would be \(1.4~\mathrm{s} \times 10 = 14~\mathrm{s}\), whereas with \(\text{VADER}_\text{raw}\), it would be \(0.022~\mathrm{s} \times 10 = 0.22~\mathrm{s}\). Even for more realistic scenarios with, for example, three sensors, \(\text{VADER}_\text{spec}\) would take approximately 4.2~seconds, while \(\text{VADER}_\text{raw}\) would process the data in under a second. This efficiency gain is crucial for real-world applications, especially when considering additional tracks or limited computing power for on-site evaluation. With trains passing every few minutes, the higher efficiency of \(\text{VADER}_\text{raw}\) makes it more suitable for practical deployment in continuous monitoring systems.

\subsection{Results of Hyperparameter Study}

\begin{figure*}
    \centering

\begin{subfigure}{0.49\textwidth}
    \centering
    \includegraphics[width=\linewidth]{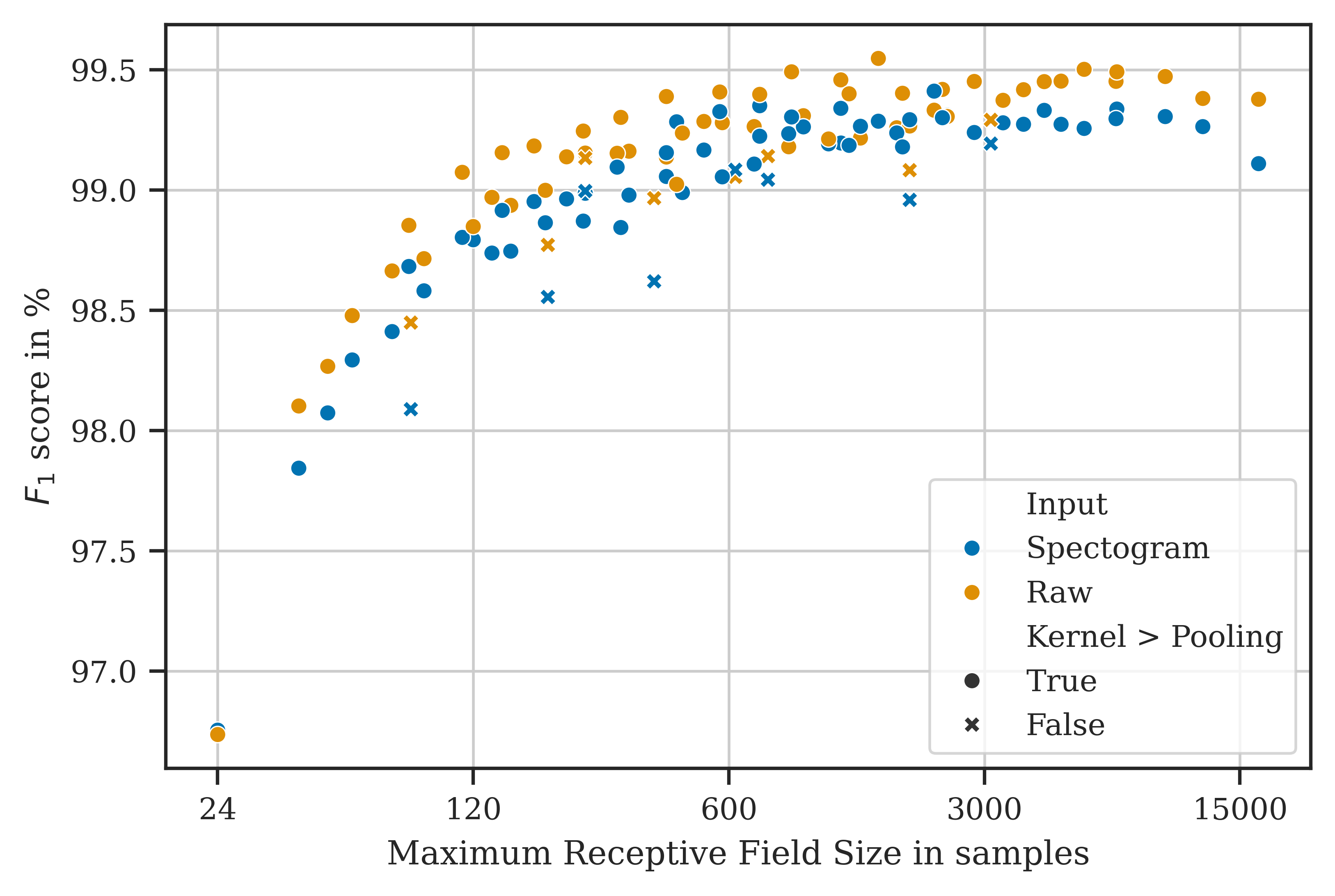}
    \caption{\(F_1\) scores for stratified scenario}
    \label{fig:hp_strat_f1}
\end{subfigure}
\hfill
\begin{subfigure}{0.49\textwidth}
    \centering
    \includegraphics[width=\linewidth]{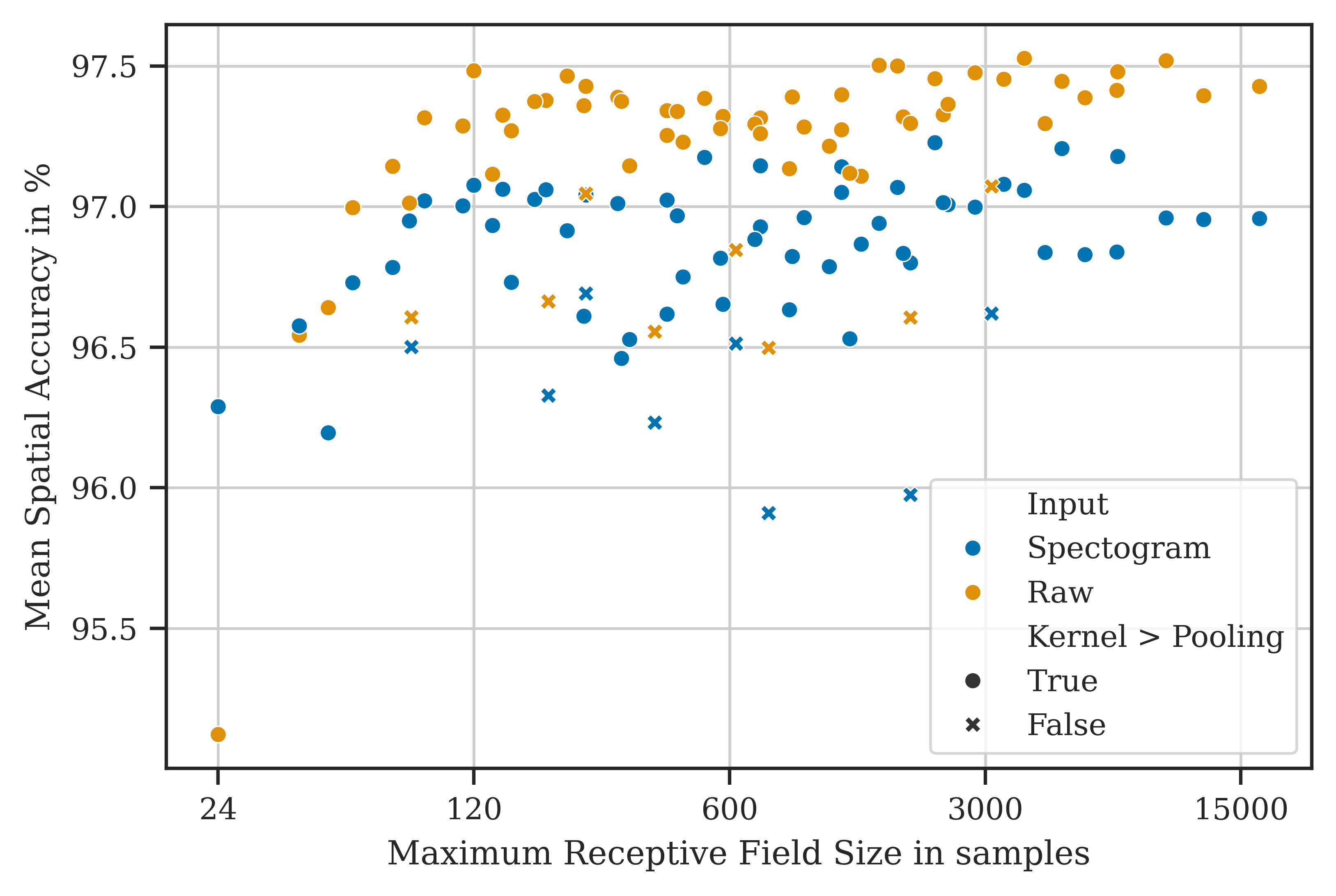}
    \caption{MSA for stratified scenario}
    \label{fig:hp_strat_msa}
\end{subfigure}
\hfill
\begin{subfigure}{0.49\textwidth}
    \centering
    \includegraphics[width=\linewidth]{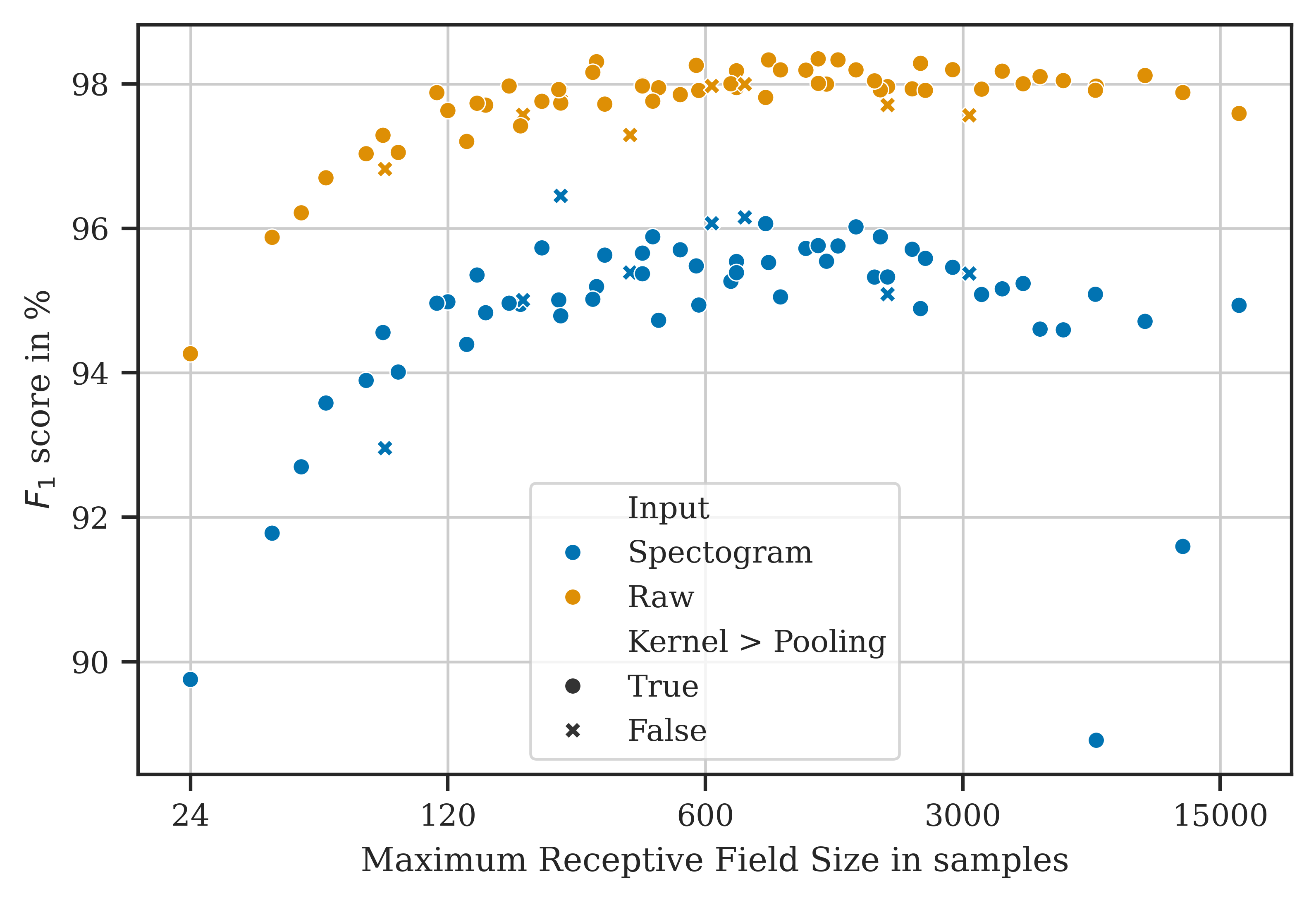}
    \caption{\(F_1\) scores for DGPS scenario}
    \label{fig:hp_dgps_f1}
\end{subfigure}
\hfill
\begin{subfigure}{0.49\textwidth}
    \centering
    \includegraphics[width=\linewidth]{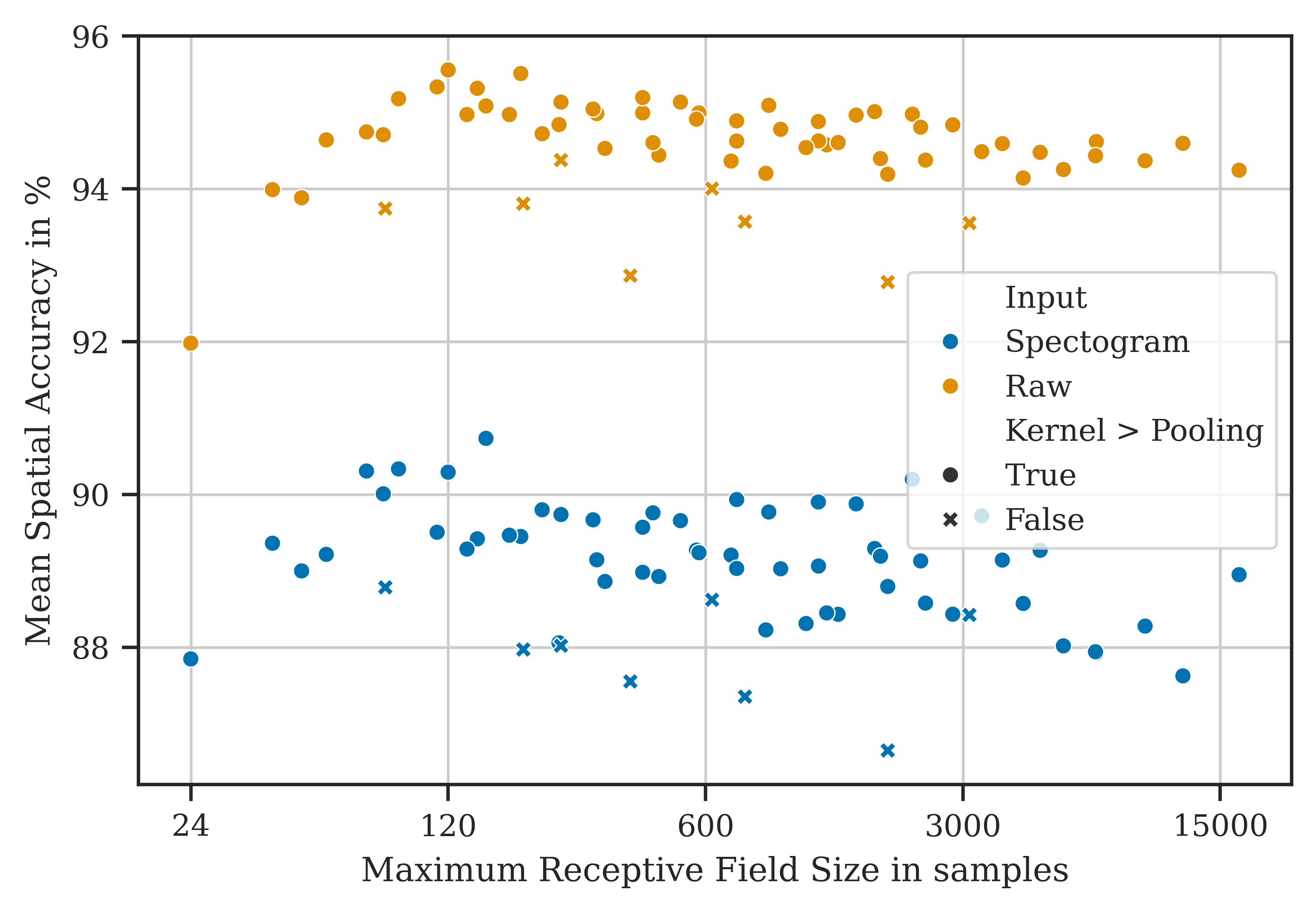}
    \caption{MSA for DGPS scenario}
    \label{fig:hp_dgps_msa}
\end{subfigure}

    \caption{Results of hyperparameter study: \(F_1\) scores and MSA as a function of the maximum receptive field, trained on stratified and DGPS scenarios. Hyperparameter combinations where the kernel size is not larger than the max pooling size are marked.}
    \label{fig:parameter_results}
\end{figure*}

The performance of both \(\text{VADER}_\text{raw}\) and \(\text{VADER}_\text{spec}\) models appears to correlate with the MRF size (Fig.~\ref{fig:parameter_results}). In the stratified scenario, this correlation is most evident for both the \(F_1\) score (Fig.~\ref{fig:hp_strat_f1}) and the MSA (Fig.~\ref{fig:hp_strat_msa}). The results of \(\text{VADER}_\text{raw}\) and \(\text{VADER}_\text{spec}\) overlap, although \(\text{VADER}_\text{raw}\) generally performs better on average for each MRF size. Notably, the MSA reaches a plateau from an MRF size of approximately 120 samples, especially when excluding hyperparameter combinations where the kernel size is not larger than the max pooling size (Fig.~\ref{fig:hp_strat_msa}).

When the kernel size is smaller than or equal to the max pooling size, the signal cannot be processed correctly in the decoder path (or upsampling path) using transposed convolution. In this context, the max pooling size acts as an upsampling factor. For example, with a max pooling size of three, the signal must be increased to three times its length during upsampling. This is typically achieved by inserting two zeros between each input value, and then applying the transposed convolution. However, if the kernel of the transposed convolution only stretches over one input value and two zeros (due to the kernel size being smaller or equal to the pooling size), the transposed convolution layer cannot effectively interpolate between input values. Once the kernel size is larger than the max pooling size, the kernel can capture multiple input values, allowing proper interpolation.

The impact of upsampling without interpolation is more pronounced for the MSA (Fig.~\ref{fig:hp_strat_msa}) than for the \(F_1\) score (Fig.~\ref{fig:hp_strat_f1}). This is because the MSA is sensitive to temporal and spatial errors, whereas the \(F_1\) score simply measures whether an axle has been detected or not, regardless of minor shifts in detection time.

In the DGPS scenario, the correlation between model performance and MRF size is less clear. The \(\text{VADER}_\text{raw}\) models perform significantly better than \(\text{VADER}_\text{spec}\), with no overlap in results. For the \(F_1\) score, there is a slight peak at an MRF size of approximately 600 samples, more evident in \(\text{VADER}_\text{spec}\) (Fig.~\ref{fig:hp_dgps_f1}). The MSA shows a clear maximum at an MRF size of around 120 samples (Fig.~\ref{fig:hp_dgps_msa}). Based on these observations, hyperparameter combinations where the kernel size is smaller than or equal to the max pooling size were excluded from further analyses.

\begin{figure*}
    \centering
    \begin{subfigure}{0.49\textwidth}
        \centering
        \includegraphics[width=\linewidth]{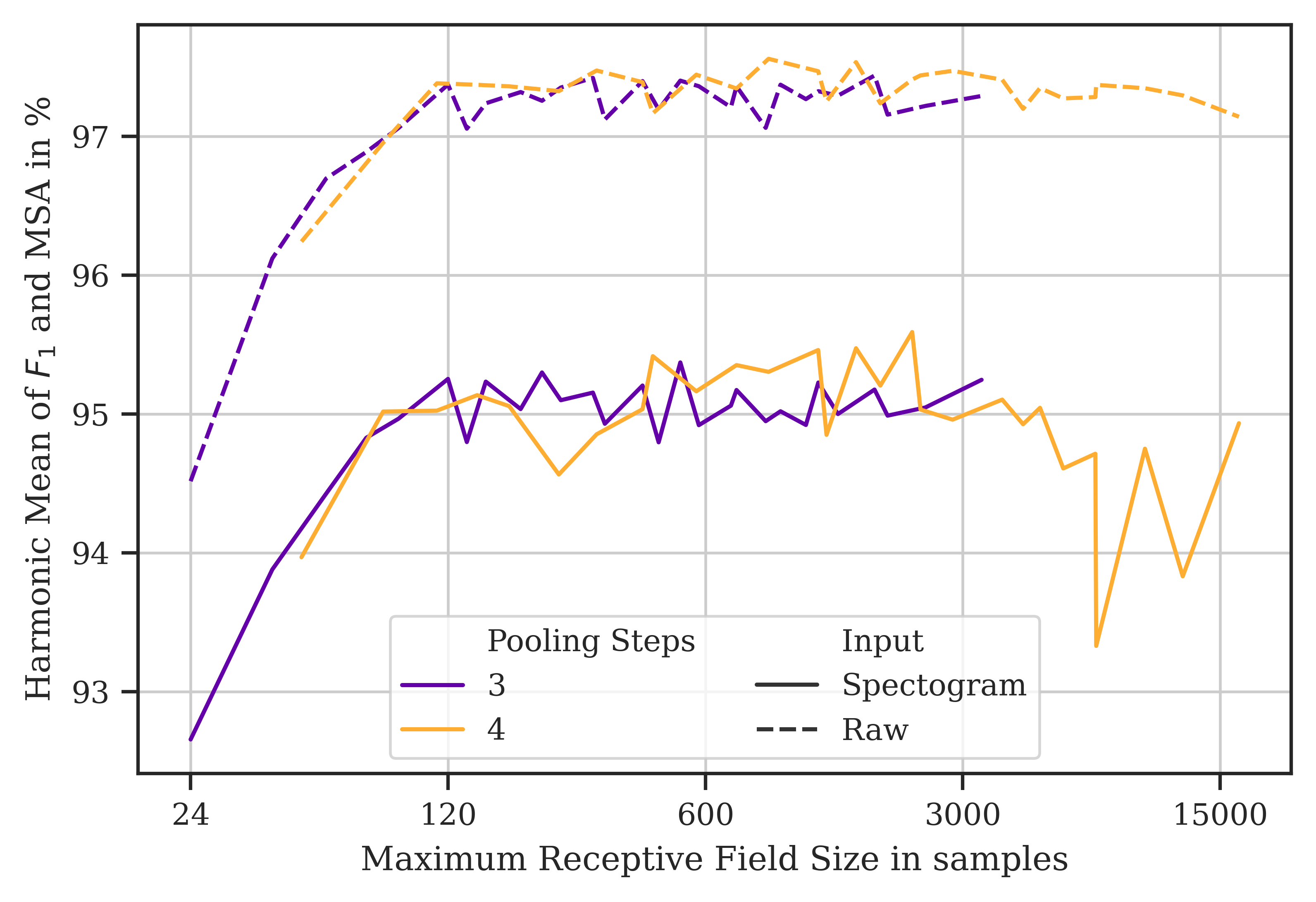}
        \caption{Impact of the number of pooling steps}
        \label{fig:pooling_steps}
    \end{subfigure}
    \hfill
    \begin{subfigure}{0.49\textwidth}
        \centering
        \includegraphics[width=\linewidth]{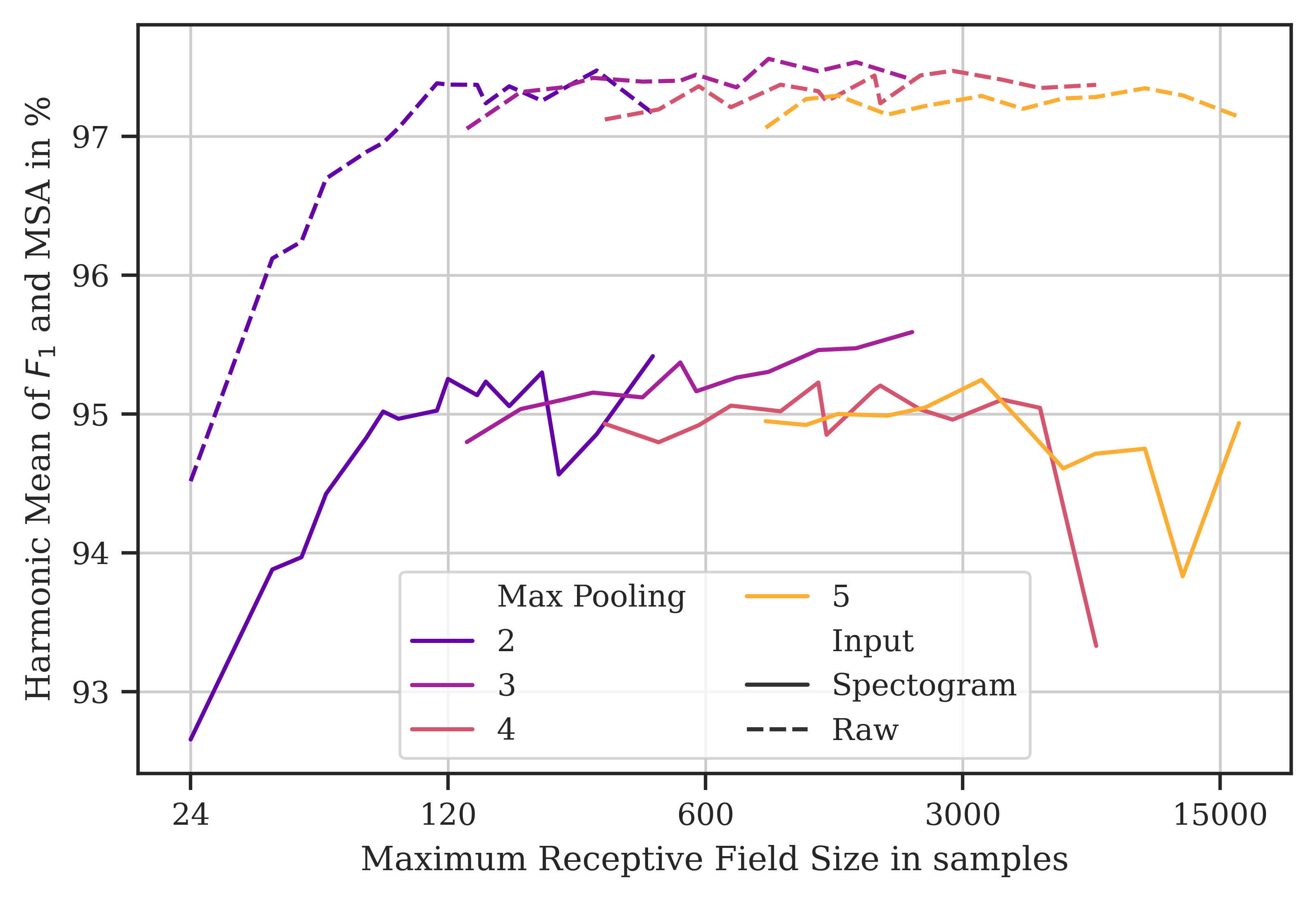}
        \caption{Impact of max pooling size}
        \label{fig:max_pooling}
    \end{subfigure}
    \hfill
    \begin{subfigure}{0.49\textwidth}
        \centering
        \includegraphics[width=\linewidth]{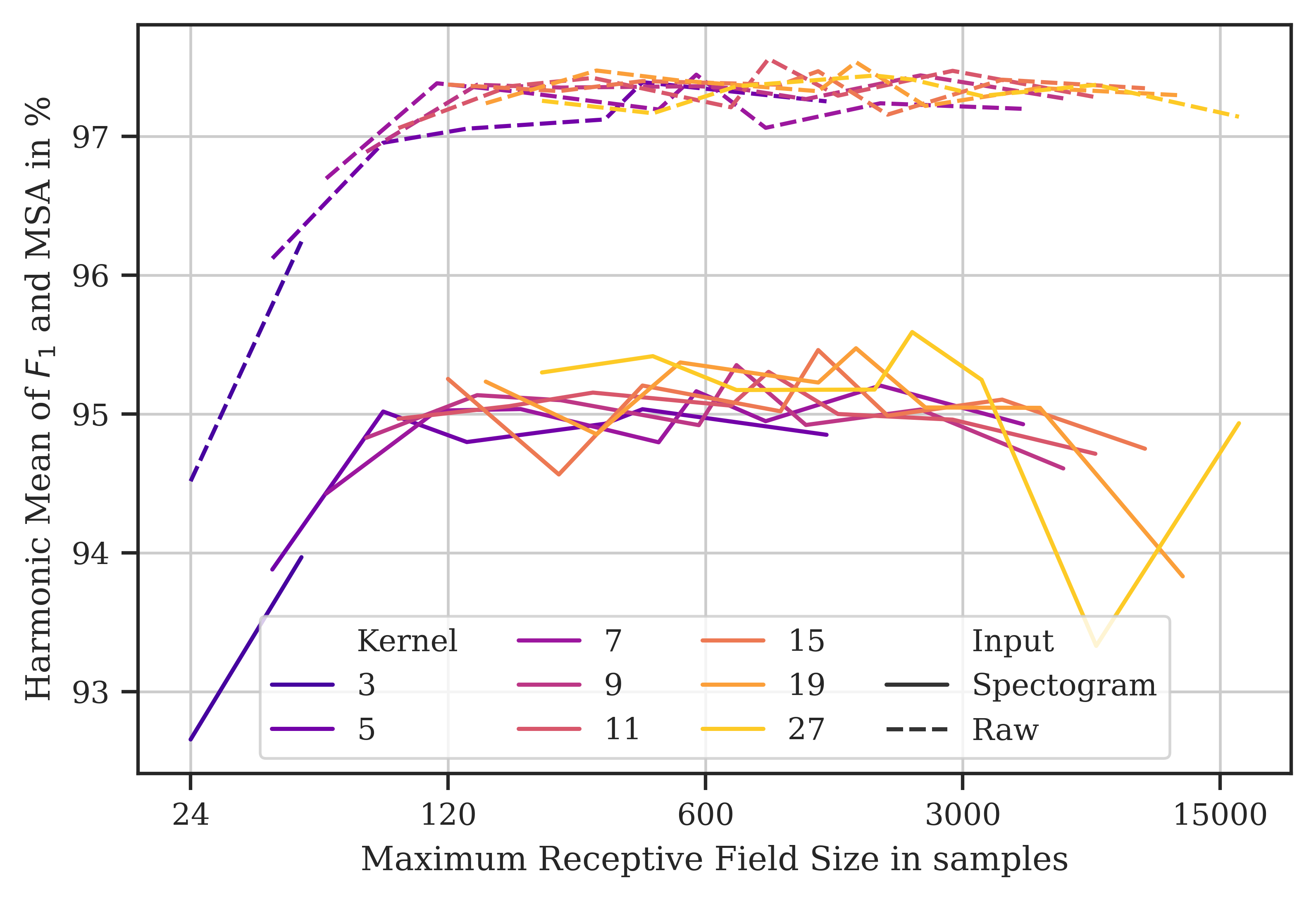}
        \caption{Impact of kernel size}
        \label{fig:kernel_size}
    \end{subfigure}

    \caption{Harmonic mean of both scenarios (stratified and DGPS) and both metrics (\(F_1\) score and MSA) as a function of maximum receptive field size. The results of all hyperparameter combinations are shown as a function of one hyperparameter.}
    \label{fig:harmonic_mean}
\end{figure*}

To select the final hyperparameters for \(\text{VADER}_\text{raw}\), we combined the results from both scenarios (stratified and DGPS) and both metrics (\(F_1\) score and MSA) by calculating the harmonic mean (HM) (Eq.~\ref{eq:hm}). This approach ensures that the chosen hyperparameters deliver strong performance across all metrics and scenarios. Overall, no substantial differences were observed between the hyperparameter combinations (Fig.~\ref{fig:harmonic_mean}). The MRF size seems to have the greatest influence on the number of pooling steps, while the results are otherwise in a similar range (Fig.~\ref{fig:pooling_steps}). There are more noticeable differences depending on the max pooling size, particularly with spectrogram inputs (Fig.~\ref{fig:max_pooling}), where pooling sizes of two and three tend to perform slightly better. No clear optimum was found for kernel sizes, but kernel sizes in the middle range (seven to 15) generally performed best (Fig.~\ref{fig:kernel_size}).

For the final evaluation, we selected two hyperparameter combinations within the optimal ranges: a max pooling size of two to three (Fig.~\ref{fig:max_pooling}), a kernel size of nine (within the optimal seven to 15 range), and an MRF size of at least 120 to 600 samples (Fig.~\ref{fig:parameter_results}). To ensure consistency between models, we fixed the number of max pooling steps at four. By varying only the max pooling sizes (two and three), we achieved MRF sizes of 144 and 729 samples, respectively. The models can thus theoretically recognize frequencies down to approximately 4.2~Hz and 0.8~Hz. This allows us to test the hypothesis that good results can be achieved when the lowest amplitude-dependent fundamental frequency of 5~Hz can be recognized by the model.

To differentiate the model names, we append the input type (raw or spec) and the max pooling size (two or three) to the VADER model name. In total, four models (\(\text{VADER}_\text{raw2}\), \(\text{VADER}_\text{raw3}\), \(\text{VADER}_\text{spec2}\), and \(\text{VADER}_\text{spec3}\)) were tested using both scenarios (stratified and DGPS) and five cross-validation splits.

\subsection{Validation Results}

\begin{figure*}
    \centering

\begin{subfigure}{0.49\textwidth}
    \centering
    \includegraphics[width=\linewidth]{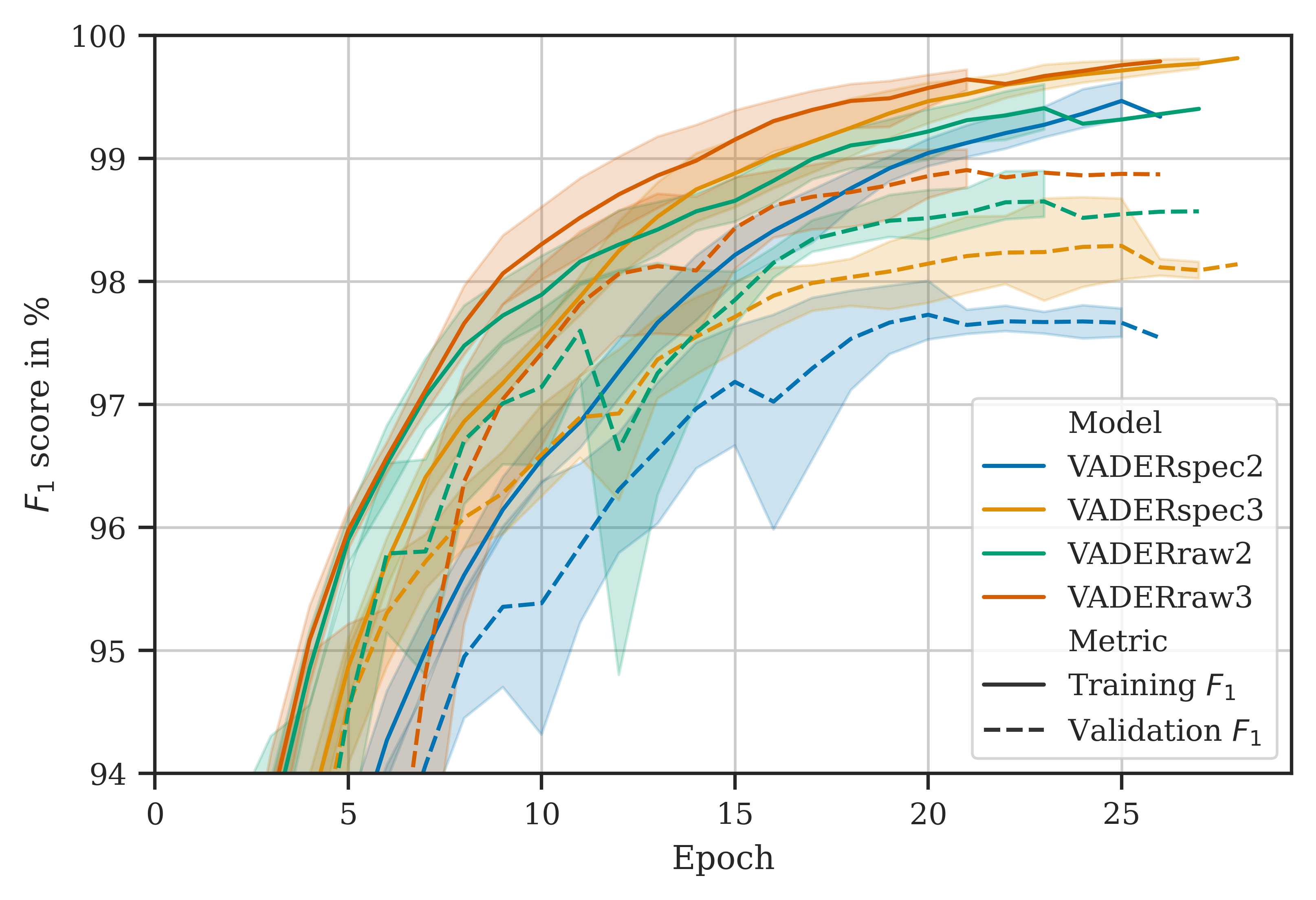}
    \caption{\(F_1\) scores in \% for DGPS scenario}
    \label{fig:f1_single}
\end{subfigure}
\hfill
\begin{subfigure}{0.49\textwidth}
    \centering
    \includegraphics[width=\linewidth]{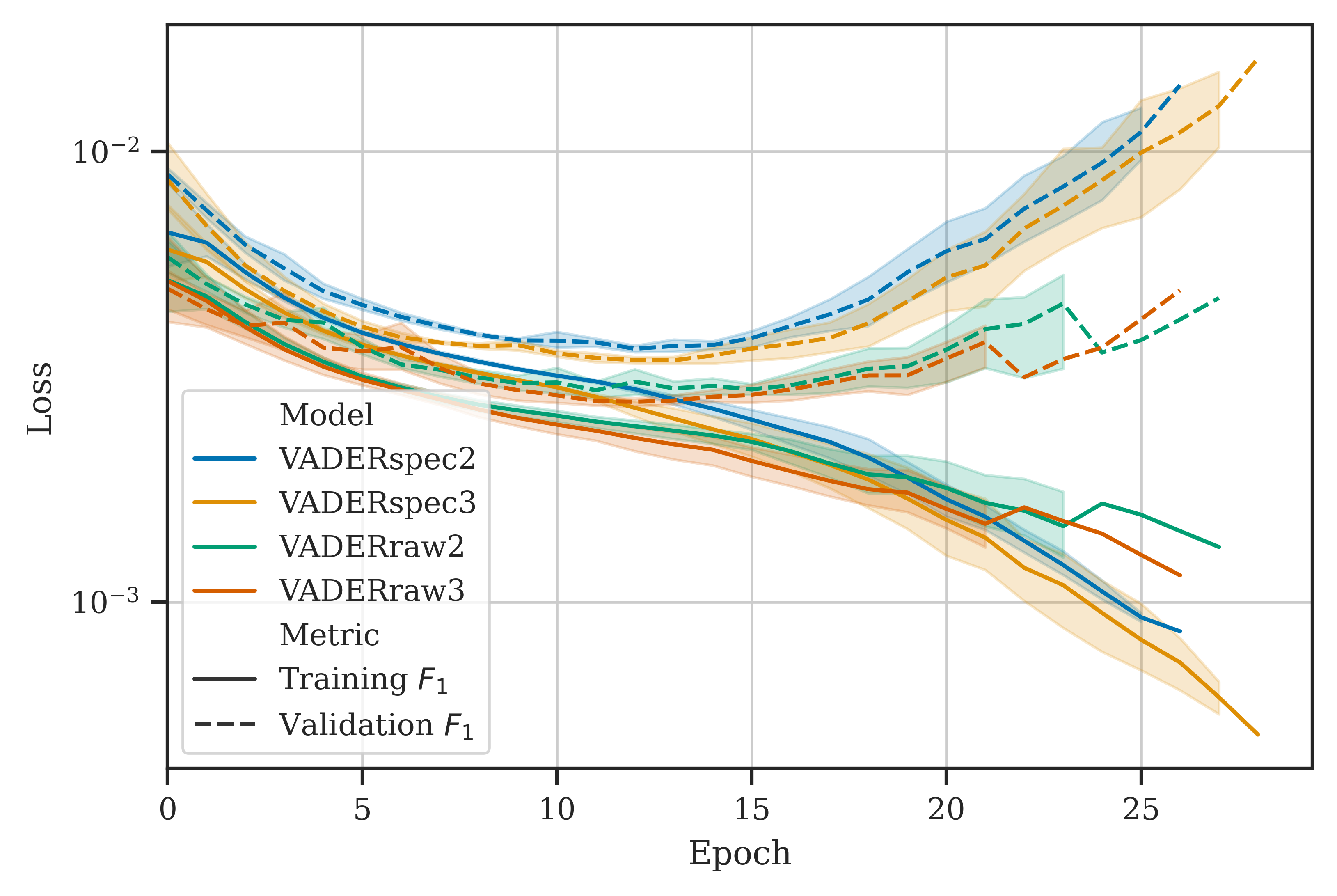}
    \caption{Binary focal loss with \(\gamma = 2.5\) for DGPS scenario}
    \label{fig:loss_single}
\end{subfigure}
\hfill
\begin{subfigure}{0.49\textwidth}
    \centering
    \includegraphics[width=\linewidth]{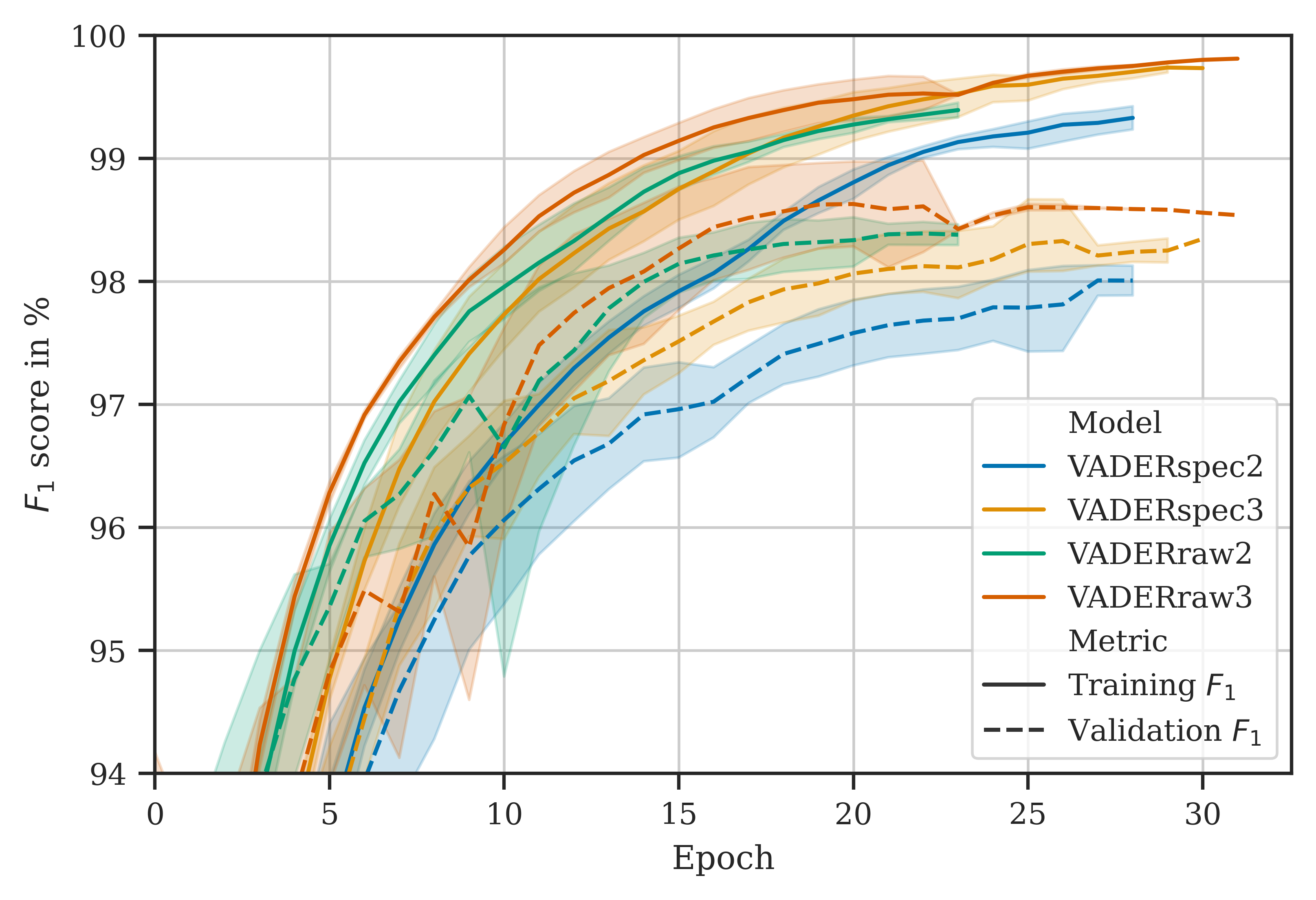}
    \caption{\(F_1\) scores in \% for stratified scenario}
    \label{fig:f1_all}
\end{subfigure}
\hfill
\begin{subfigure}{0.49\textwidth}
    \centering
    \includegraphics[width=\linewidth]{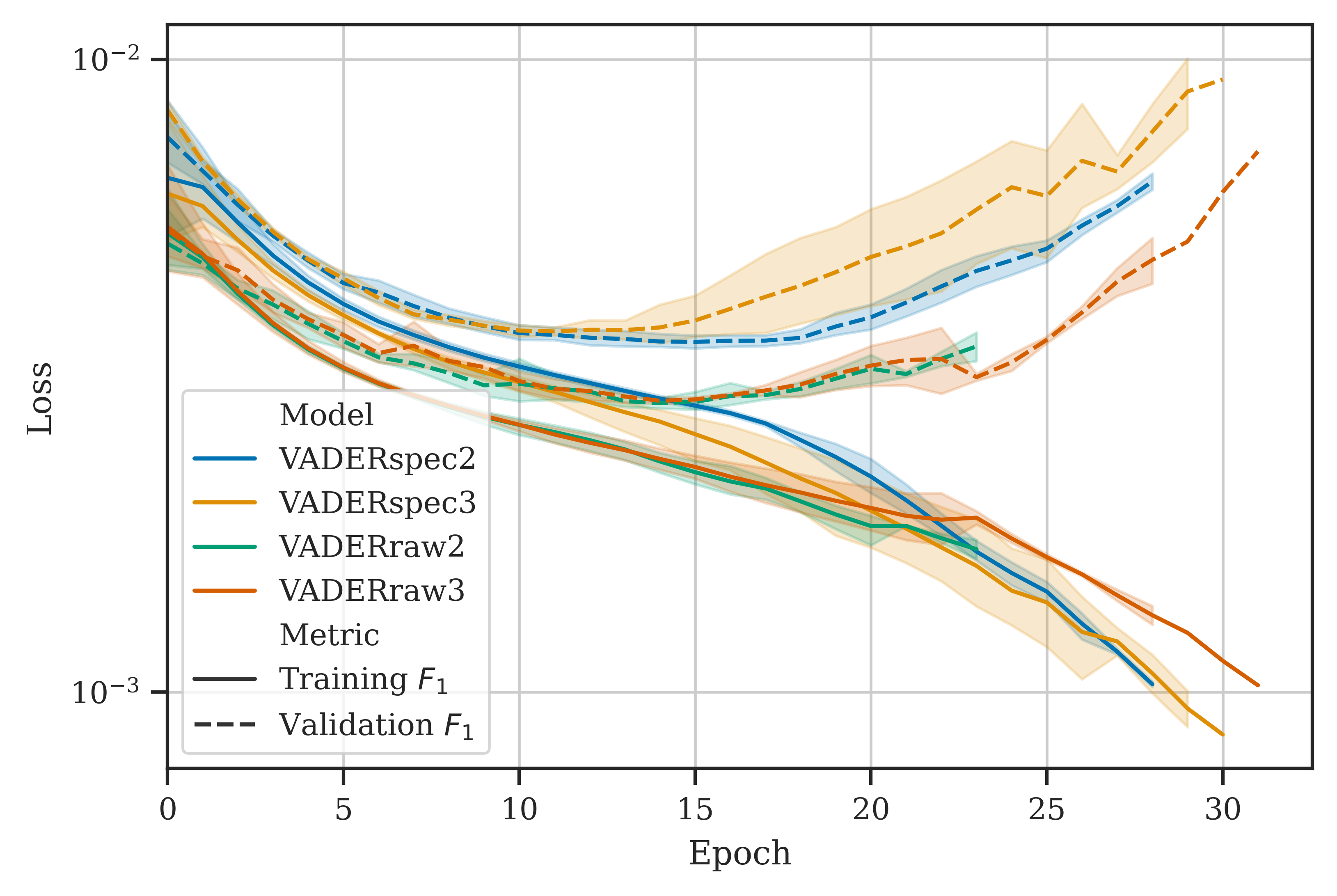}
    \caption{Binary focal loss with \(\gamma = 2.5\) for stratified scenario}
    \label{fig:loss_all}
\end{subfigure}

    \caption{Training and validation history per epoch with 95~\% confidence interval, determined via the five cross-validation splits}
    \label{fig:training_results}
\end{figure*}

To assess the generalization ability of the models, we examined both the loss and the \(F_1\) score over training epochs (Fig.~\ref{fig:training_results}). The loss curves indicate that all models exhibit typical learning patterns, with loss decreasing as training progresses. However, signs of overfitting are evident where the validation loss begins to increase after around 15 epochs, despite the continued decrease in training loss (Fig.~\ref{fig:loss_single},~\ref{fig:loss_all}). This suggests that the models start to memorize the training data rather than generalizing to new data. This effect is more pronounced in the models with spectrogram inputs, where even the minimum validation loss is significantly higher than in the models with raw data inputs.

To mitigate overfitting and maximize generalization capability, we employed early stopping and restored the weights corresponding to the best validation \(F_1\) score for use on the test set. The \(F_1\) score curves demonstrate that while all models improve in classification performance over time, those based on raw data (\(\text{VADER}_\text{raw2}\) and \(\text{VADER}_\text{raw3}\)) achieve higher final \(F_1\) scores compared to the models trained on spectrogram data (Fig.~\ref{fig:f1_single},~\ref{fig:f1_all}). The closer alignment between training and validation \(F_1\) scores in the raw data models further indicates better generalization.

Overall, these results suggest that preprocessing steps, such as converting raw data into spectrograms, may decrease model performance and increase the risk of overfitting. Therefore, \(\text{VADER}_\text{raw2}\) and \(\text{VADER}_\text{raw3}\) may be more reliable for generalization to unseen data.

While the loss is calculated per sample and thus only counts exact hits (the precise time an axle passes the sensor), the \(F_1\) score is calculated per train axle. This means that for all models and scenarios, the frequency of exact hits may decrease after a certain point in training due to overfitting, but the models may still recognize more axles overall. Depending on the specific requirements of the application, either the loss or the \(F_1\) score can be used as the criterion for terminating training. Since small spatial deviations in axle detection are usually negligible, we used the \(F_1\) score as the criterion for this study.

\subsection{Test Results}

For the test results, we used the learned parameters (weights) from the epoch with the best validation \(F_1\) score. The models were then evaluated on the previously unseen test set. This procedure was repeated for all splits, scenarios (stratified and DGPS), and thresholds (200~cm and 37~cm) for each of the four models.

\begin{figure*}
    \centering
\begin{subfigure}{0.49\textwidth}
    \centering
    \includegraphics[width=\linewidth]{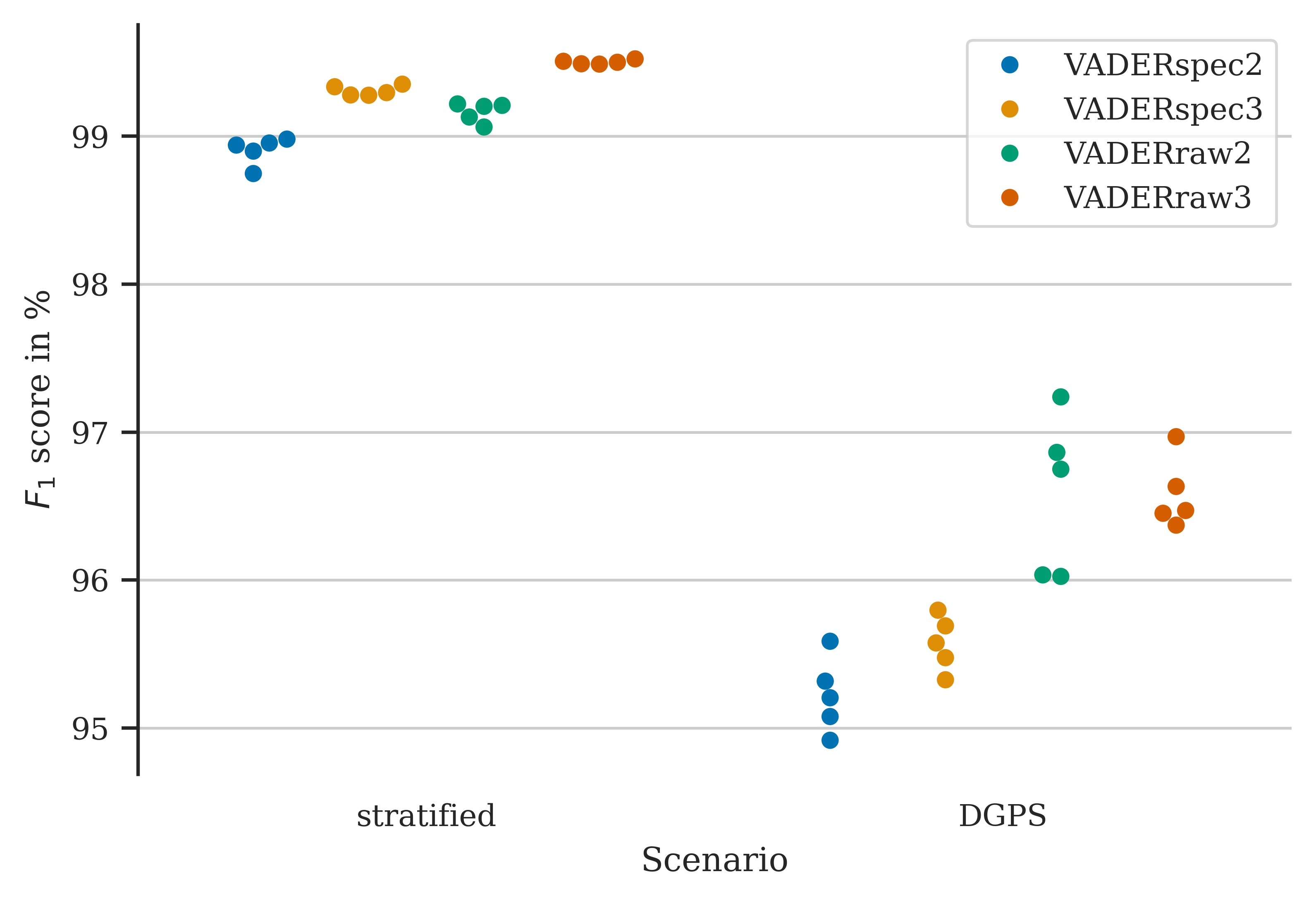}
    \caption{\(F_1\) score}
    \label{fig:f1_2}
\end{subfigure}
\hfill
\begin{subfigure}{0.49\textwidth}
    \centering
    \includegraphics[width=\linewidth]{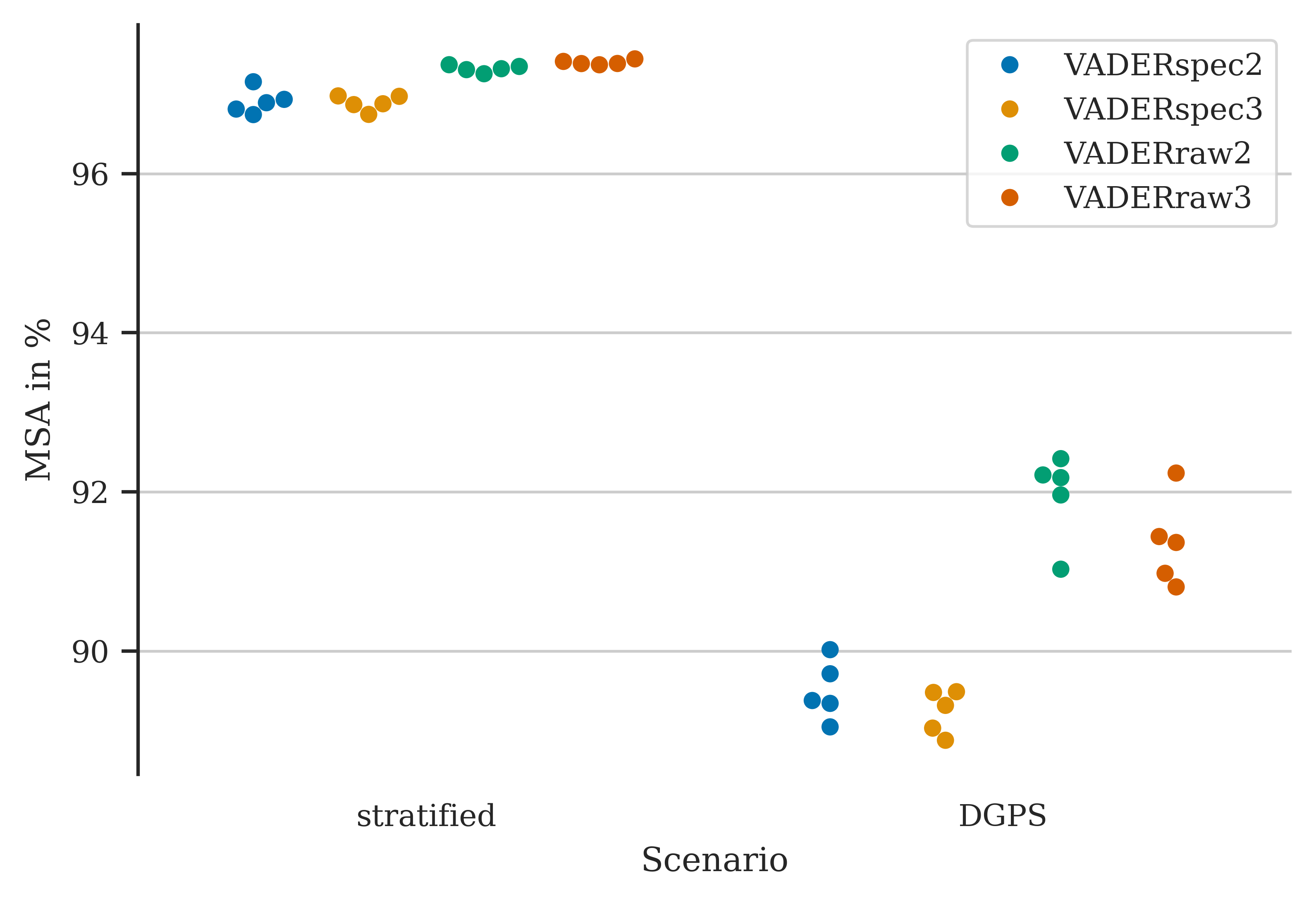}
    \caption{MSA}
    \label{fig:spatial_2}
\end{subfigure}

    \caption{\(F_1\) score and MSA of the models for the test sets of both scenarios with a classification threshold of 200~cm. Each data point represents the average results per split.}
    \label{fig:boxplot}
\end{figure*}

Overall, the accuracy of all models is considerably better for the stratified scenario compared to the DGPS scenario (Fig.~\ref{fig:f1_2},~\ref{fig:spatial_2}). The variance of the results is comparable across all models and, as expected, generally higher for the more challenging DGPS scenario. The \(\text{VADER}_\text{raw}\) models achieve significantly higher accuracy in the DGPS scenario (Fig.~\ref{fig:f1_2},~\ref{fig:spatial_2}). In the stratified scenario, the \(\text{VADER}_\text{raw}\) models also perform slightly better in terms of MSA (Fig.~\ref{fig:spatial_2}), but the \(F_1\) score results are mixed (Fig.~\ref{fig:f1_2}). In this case, the models with a max pooling size of three (and thus a larger MRF size) perform better than their counterparts with a max pooling size of two.

\(\text{VADER}_\text{raw2}\) achieves the best results in all metrics for the DGPS splits, while \(\text{VADER}_\text{raw3}\) achieves the best results for the stratified splits (Table~\ref{tab:comparison}). Since \(\text{VADER}_\text{raw3}\) has a significantly larger MRF, it may learn entire train axle configurations, whereas \(\text{VADER}_\text{raw2}\) may focus on individual axles or bogies. Learning entire train axle configurations could be advantageous for a representative dataset (stratified scenario), but may increase the risk of overfitting on the training data in non-representative data scenarios (DGPS scenario). Generally, the \(\text{VADER}_\text{raw}\) models outperform the \(\text{VADER}_\text{spec}\) models.

The only other comparable method for axle detection using arbitrarily placed acceleration sensors (and thus applicable to our dataset) is the VAD from \citet{Lorenzen2022VirtualAD}. VAD was evaluated in a scenario comparable to our stratified scenario, achieving a \(\overline{\Delta s}\) of 10.3~cm and an \(F_1\) score of 95.4~\% and 91.5~\% with threshold values of 200~cm and 37~cm, respectively. As expected, VAD performs significantly worse than the optimized \(\text{VADER}_\text{spec}\) models (Table~\ref{tab:comparison}).

\begin{table}[h!]
    \centering
    
    \begin{tabular}{lllll}
    \toprule
    Scenario &       Model &   \(\overline{\Delta s}\) [cm] &  \multicolumn{2}{c}{\(F_1\) Score} \\
     & & & 200~cm & 37~cm \\
    \midrule
    DGPS &  \(\text{VADER}_\text{spec2}\) & 21.0  &      95.2~\% &     78.2~\% \\
     &  \(\text{VADER}_\text{spec3}\) & 21.5  &      95.6~\% &     78.2~\% \\
     &     \(\text{VADER}_\text{raw2}\) & \textbf{16.1}  &      \textbf{96.6}~\% &     \textbf{83.4}~\% \\
     &     \(\text{VADER}_\text{raw3}\) & 17.3  &      \textbf{96.6}~\% &     82.7~\% \\
    \midrule
    Stratified &  \(\text{VADER}_\text{spec2}\) & 6.19  &      98.9~\% &     95.7~\% \\
     &  \(\text{VADER}_\text{spec3}\) & 6.23  &      99.3~\% &     95.8~\% \\
     &     \(\text{VADER}_\text{raw2}\) & 5.36  &      99.2~\% &     96.5~\% \\
     &     \(\text{VADER}_\text{raw3}\) & \textbf{5.21}  &      \textbf{99.5}~\% &     \textbf{96.8}~\% \\
    \bottomrule
    \end{tabular}
    
    \caption{\(F_1\) score and \(\overline{\Delta s}\) calculated for all samples of the corresponding splits and scenarios}
    \label{tab:comparison}
\end{table}

\subsection{Sensor-Dependent Test Results}

\begin{figure*}
\begin{subfigure}{0.49\textwidth}
    \centering
    \includegraphics[width=\linewidth]{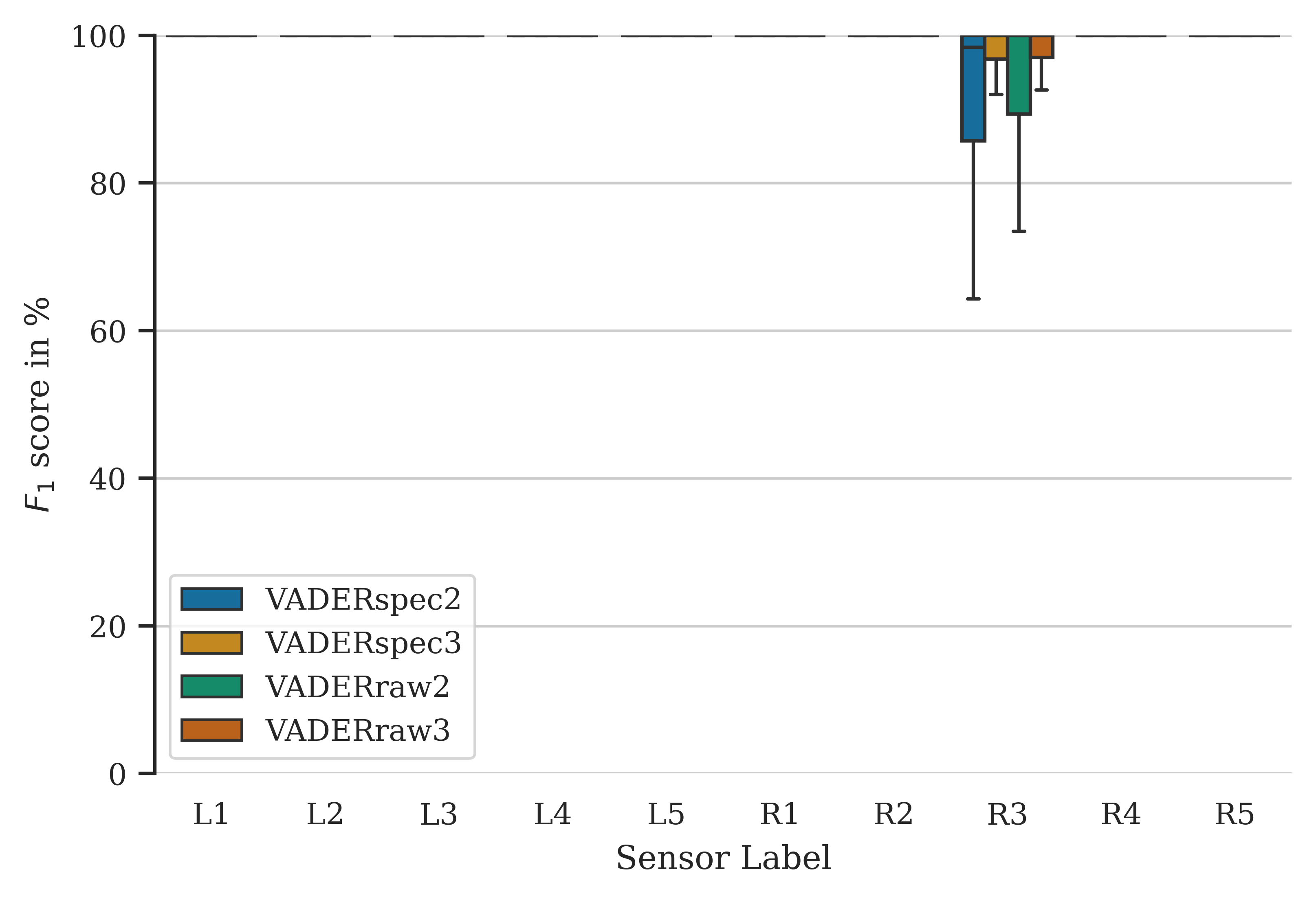}
    \caption{\(F_1\) score with a threshold of 200~cm for each sensor, stratified scenario}
    \label{fig:sensor_all_200}
\end{subfigure}
\hfill
\begin{subfigure}{0.49\textwidth}
    \centering
    \includegraphics[width=\linewidth]{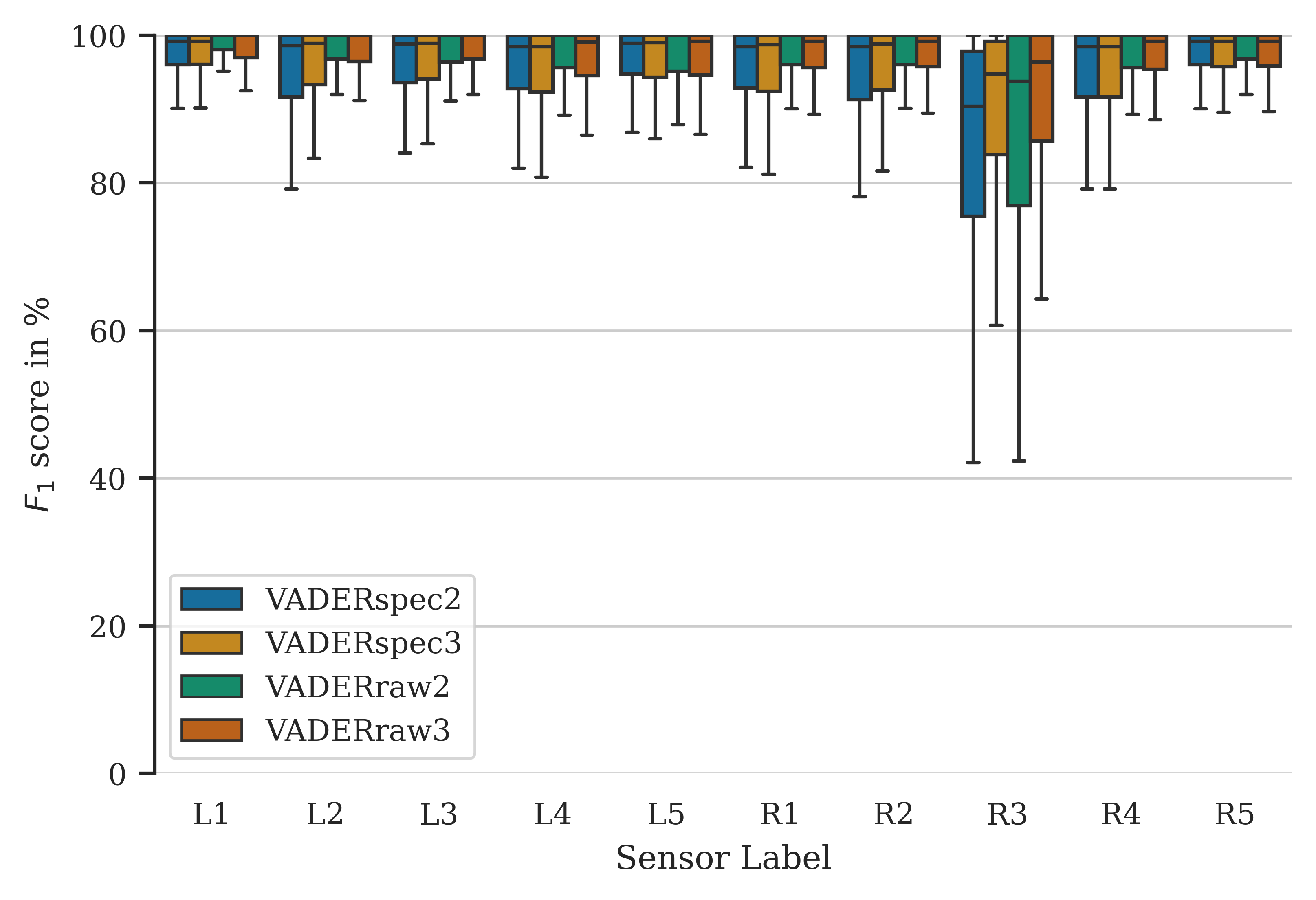}
    \caption{\(F_1\) score with a threshold of 200~cm for each sensor, DGPS scenario}
    \label{fig:sensor_single_200}
\end{subfigure}
\hfill
\begin{subfigure}{0.49\textwidth}
    \centering
    \includegraphics[width=\linewidth]{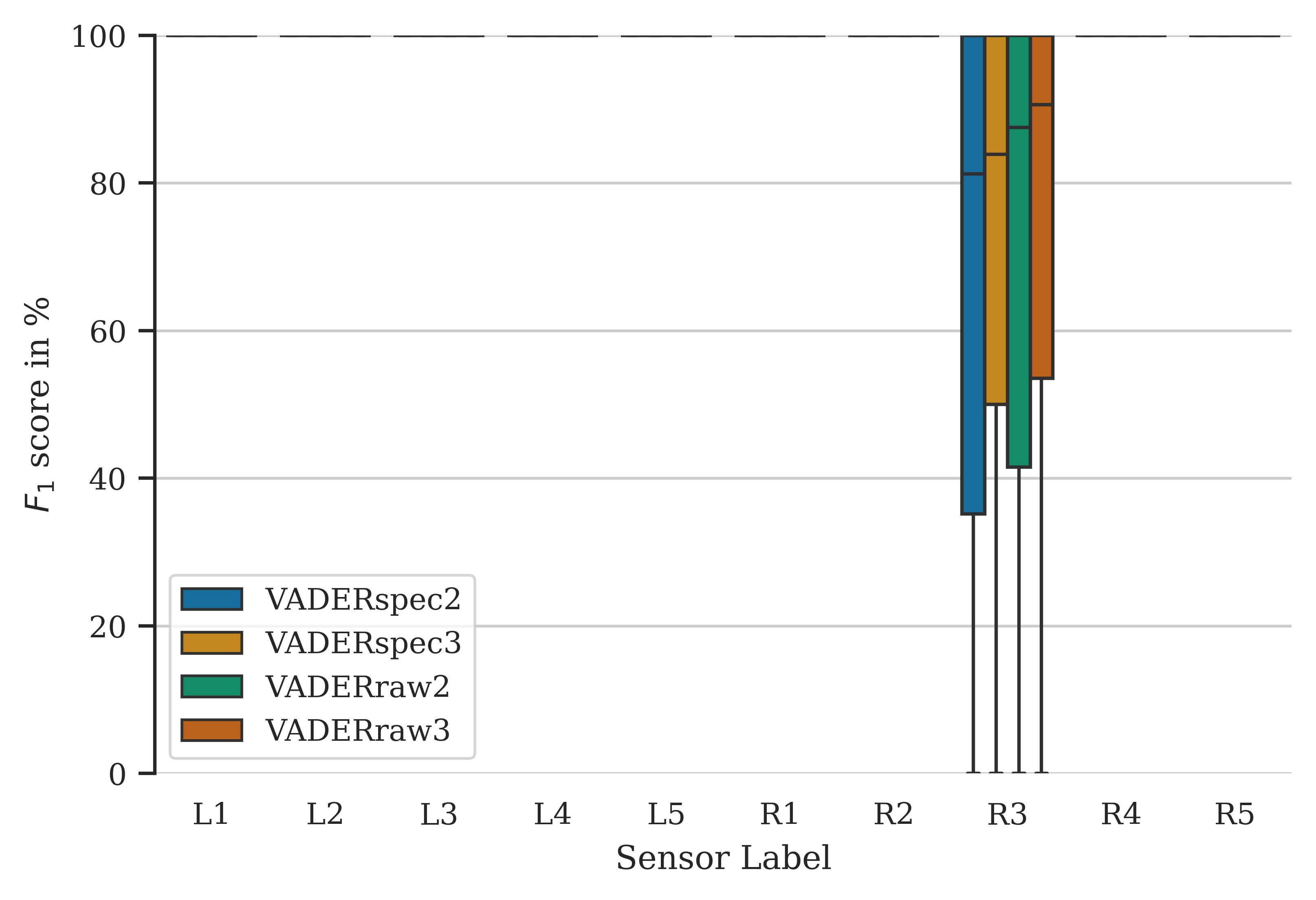}
    \caption{\(F_1\) score with a threshold of 37~cm for each sensor, stratified scenario}
    \label{fig:sensor_all_37}
\end{subfigure}
\hfill
\begin{subfigure}{0.49\textwidth}
    \centering
    \includegraphics[width=\linewidth]{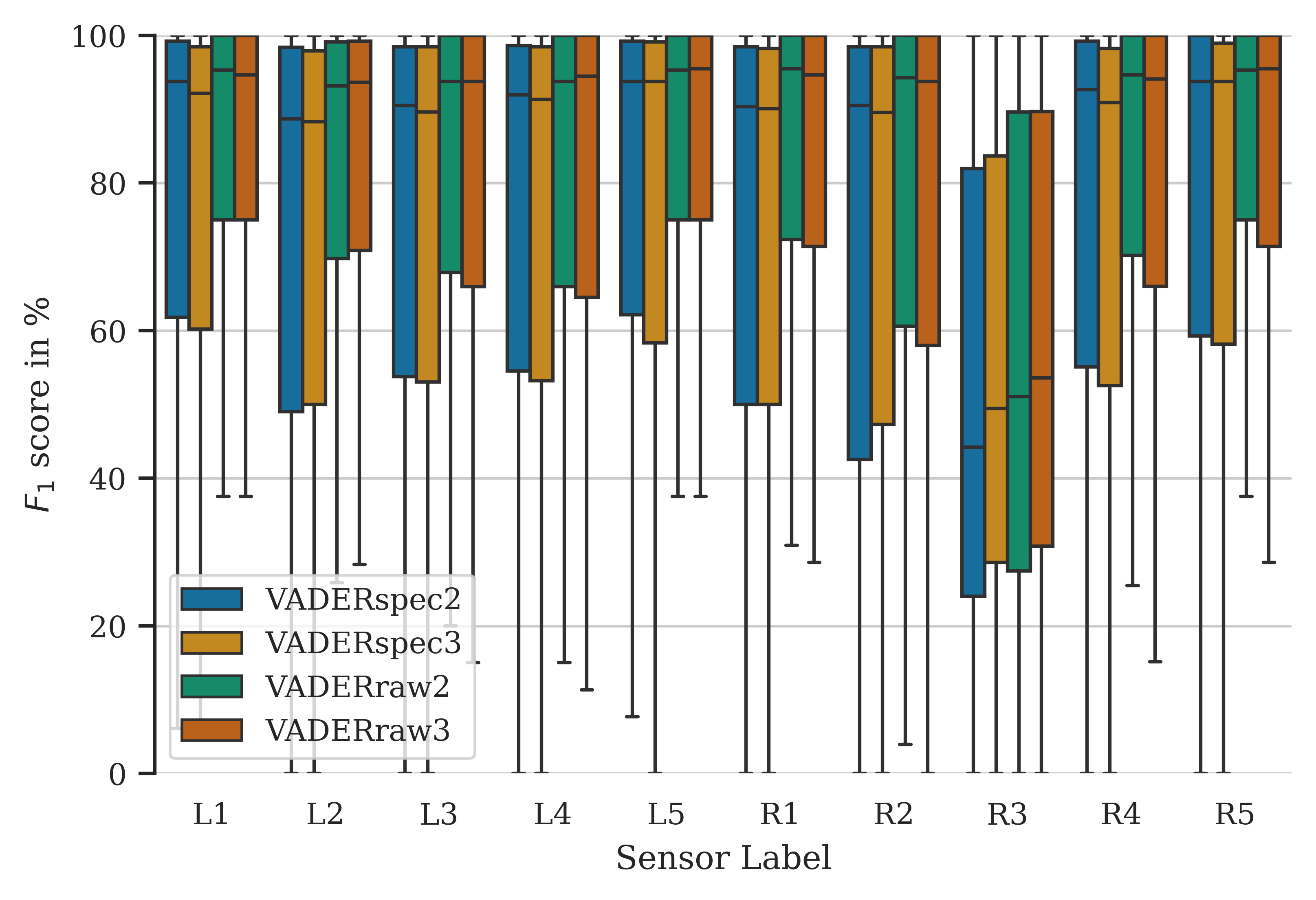}
    \caption{\(F_1\) score with a threshold of 37~cm for each sensor, DGPS scenario}
    \label{fig:sensor_single_37}
\end{subfigure}
\hfill
\begin{subfigure}{0.49\textwidth}
    \centering
    \includegraphics[width=\linewidth]{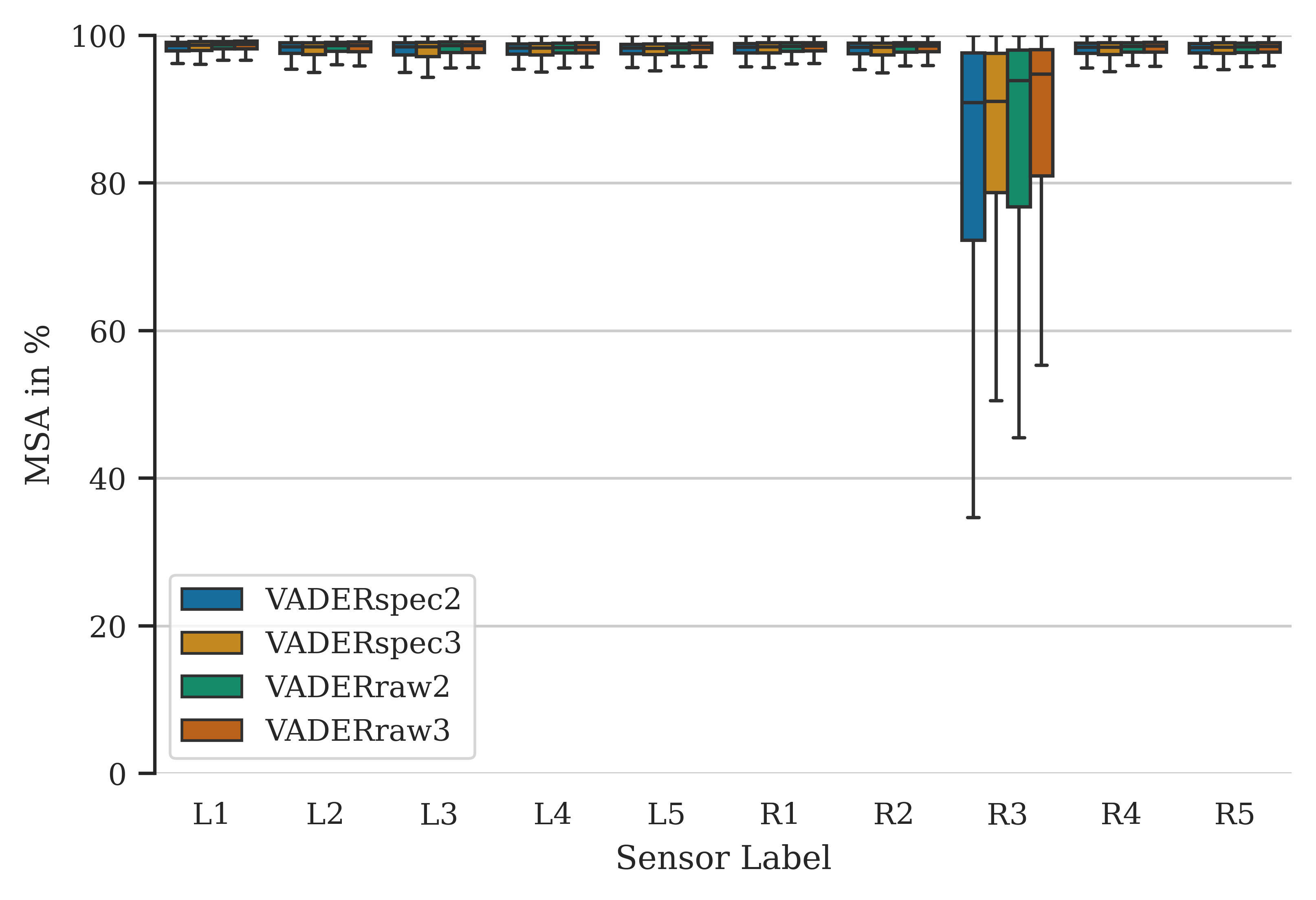}
    \caption{MSA for each sensor, stratified scenario}
    \label{fig:sensor_all_2_SE}
\end{subfigure}
\hfill
\begin{subfigure}{0.49\textwidth}
    \centering
    \includegraphics[width=\linewidth]{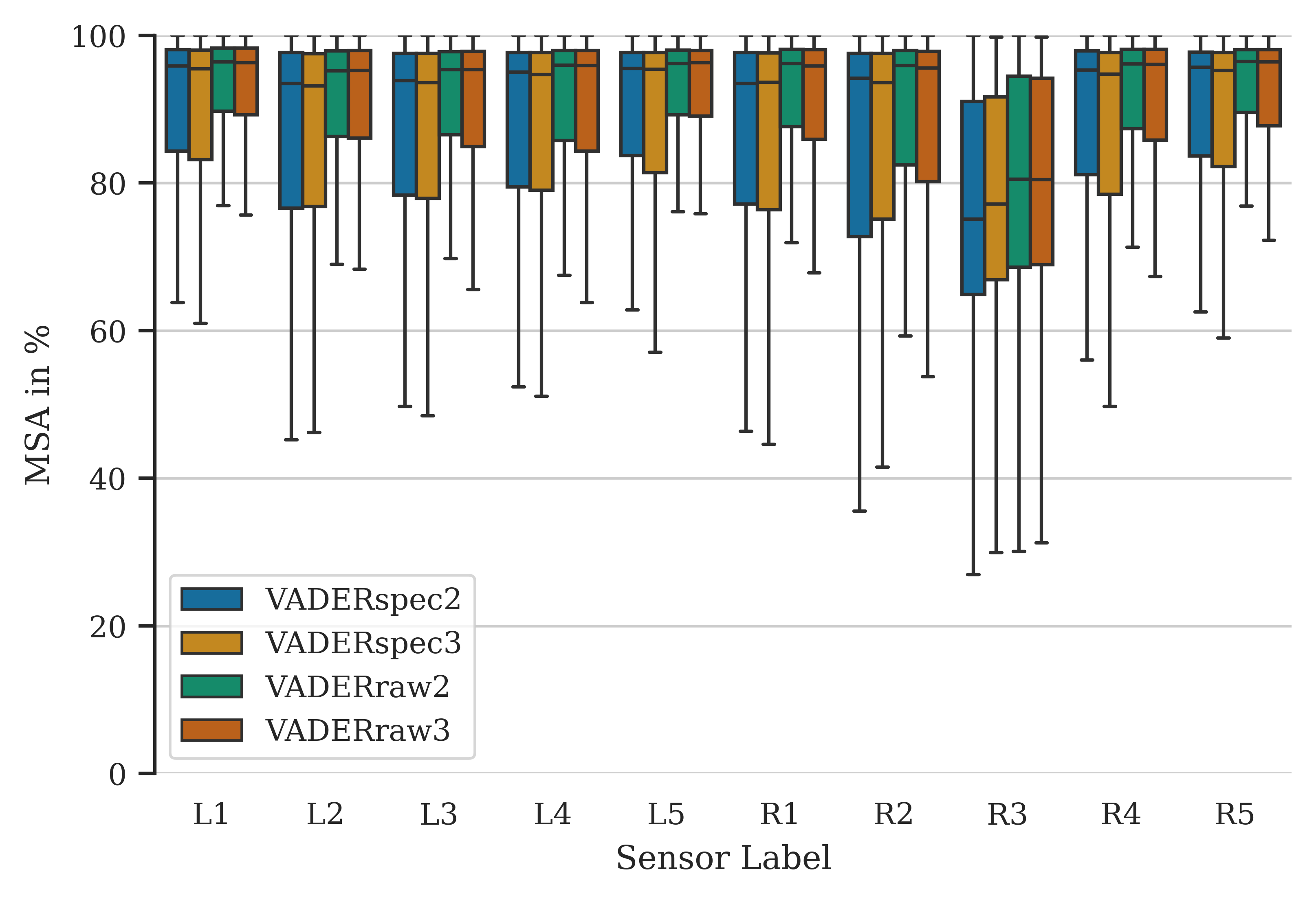}
    \caption{MSA for each sensor, DGPS scenario}
    \label{fig:sensor_single_2_SE}
\end{subfigure}
    \caption{Model performance metrics evaluated per sensor}
    \label{fig:metrics_per_sensor}
\end{figure*}

Previous observations indicated that one of the sensors (R3) was degraded during the measurement campaign \citep{Lorenzen2022VirtualAD}. To investigate the influence of sensor degradation on model performance, we examined the \(F_1\) score and MSA for each sensor, across both scenarios and threshold values (Figs.~\ref{fig:sensor_all_200} to \ref{fig:sensor_single_2_SE}). Generally, \(\text{VADER}_\text{raw}\) outperforms \(\text{VADER}_\text{spec}\) across all quartiles in the plots, with few exceptions.

Apart from the degraded sensor R3, the results for the other sensors are very similar (Fig.~\ref{fig:metrics_per_sensor}). No systematic differences are observable based on sensor location (e.g., near the support or at the bridge center). VADER was specifically developed so that each sensor operates independently as an axle detector. Since no significant differences were found based on sensor position, and because sensors can generally be attached to the longitudinal structure of bridges, the method is potentially applicable to various bridge types. However, the effects of other parameters such as bridge stiffness and ballast thickness on the results remain unclear.

It is evident that worse results are achieved on sensor R3 (Fig.~\ref{fig:metrics_per_sensor}). This difference is most pronounced in the stratified splits, where the lower quartile for most sensors reaches 100~\%, but results for sensor R3 show greater variability. In the DGPS splits, results are more scattered for all sensors, but significant differences are still observed, particularly in the median values. For sensor R3, \(\text{VADER}_\text{raw3}\) often performs better than the other models.

When we reevaluate the results excluding sensor R3, \(\text{VADER}_\text{raw2}\) achieves the best performance across all metrics and scenarios (Table~\ref{tab:comparison_wo_r3}). At a threshold of 200~cm, 99.9~\% of the axles are recognized with a \(\overline{\Delta s}\) of only 3.69~cm. This suggests that the MRF rule, as applied in \(\text{VADER}_\text{raw2}\), effectively determines the optimal MRF size for fully functioning sensors. In cases of sensor degradation, larger MRFs may offer some advantages (Table~\ref{tab:comparison}), although the differences depend on the scenario.

We consider the hypothesis—that good results can be achieved when the largest object (lowest frequency of interest) is equal to or smaller than the MRF of a model (i.e., the MRF rule)—to be confirmed. Further investigation is required to determine if an MRF size slightly exceeding the largest object of interest is generally optimal, and whether significantly larger MRFs lead to overfitting. If validated, the MRF rule could significantly reduce the hyperparameter space for various types of unstructured data and potentially replace extensive hyperparameter tuning.

For this purpose, the MRF rule must be confirmed on other datasets (e.g., earthquakes or speech), data types (e.g., images or videos), model architectures (e.g., EfficientNet or Feature Pyramid Network), and problem settings (e.g., classification or object detection). Currently, the MRF rule can be effectively used if the largest object size of interest can be determined with reasonable effort. Techniques like pyramidal feature hierarchies, as described in \citet{fpn}, could be utilized to automatically analyze at which MRF size the model performance no longer improves.

\begin{table}
    \centering
    
    \begin{tabular}{lllll}
    \toprule
    Scenario &       Model &   \(\overline{\Delta s}\) &  \multicolumn{2}{c}{\(F_1\) Score} \\
     & & & 200~cm & 37~cm \\
    \midrule
    DGPS &  \(\text{VADER}_\text{spec2}\) & 19.2~cm &      96.2~\% &     80.7~\% \\
     &     \(\text{VADER}_\text{spec3}\) & 19.7~cm &      96.2~\% &     80.4~\% \\
     &     \(\text{VADER}_\text{raw2}\) & \textbf{14.5~cm} &      \textbf{97.5~\%} &     \textbf{85.8~\%} \\
     &     \(\text{VADER}_\text{raw3}\) & 15.6~cm &      97.1~\% &     85.0~\% \\
    \midrule
    Stratified &  \(\text{VADER}_\text{spec2}\) &    4.30~cm &      99.8~\% &     98.5~\% \\
     &  \(\text{VADER}_\text{spec3}\) &    4.55~cm &      99.8~\% &     98.1~\% \\
     &     \(\text{VADER}_\text{raw2}\) & \textbf{3.69~cm} &     \textbf{99.9~\%} &     \textbf{99.1~\%} \\
     &     \(\text{VADER}_\text{raw3}\) & 3.71~cm &      \textbf{99.9~\%} &     98.9~\% \\
    \bottomrule
    \end{tabular}

    \caption{\(F_1\) score and \(\overline{\Delta s}\) calculated for all samples of the corresponding splits and scenarios without sensor R3.}
    \label{tab:comparison_wo_r3}
\end{table}

\section{Conclusion}
\label{S:4}

This study presents significant advancements in the real-time detection of train axles using bridge-mounted sensors. By employing raw acceleration data instead of spectrograms, our proposed method, \(\text{VADER}_\text{raw}\), achieves substantial improvements in computational efficiency and memory usage while maintaining or even enhancing detection accuracy. Comparable results were achieved with each of the nine functioning sensors, regardless of their positions on the bridge. This makes \(\text{VADER}_\text{raw}\) the first method capable of real-time detection of train axles with arbitrarily installed sensors, significantly improving the practicality and flexibility of railway bridge monitoring systems.

Furthermore, we propose a novel MRF rule for optimizing CNNs based on the largest object of interest in potentially arbitrary unstructured data. This rule allows the hyperparameter space to be restricted and may even replace extensive hyperparameter tuning, streamlining the development of efficient deep learning models.

The key findings of the study are as follows:

\begin{itemize}
    \item The proposed MRF rule effectively narrows the hyperparameter space and provides pre-training insights into underfitting and potential overfitting risks, potentially eliminating the need for extensive hyperparameter tuning.
    \item Individual hyperparameters alone have minimal effect on the results; instead, the combination of hyperparameters, represented by the MRF size, is crucial for optimal model performance.
    \item \(\text{VADER}_\text{raw}\) processes data 65 times faster than spectrogram-based methods, using only 1~\% of the memory, making real-time detection feasible.
    \item Sensors can be placed anywhere on the bridge and do not need to be near supports or crossbeams, making \(\text{VADER}_\text{raw}\) potentially applicable to all bridge types and sensor configurations.
    \item \(\text{VADER}_\text{raw}\) generalises well even when trained on a non-representative dataset, achieving up to 99.9~\% axle detection accuracy with minimal spatial error across diverse train types and conditions.
    \item The \(\text{VADER}_\text{raw}\) models demonstrated robustness against sensor degradation and performed consistently across different scenarios, underscoring their reliability and potential for deployment in diverse environments.
\end{itemize}

These advancements hold the promise of significantly improving the safety and reliability of railway infrastructure monitoring. By enabling accurate and efficient axle detection without the need for precise sensor placement or extensive preprocessing, \(\text{VADER}_\text{raw}\) can facilitate more widespread adoption of real-time monitoring systems, potentially leading to earlier detection of structural issues and improved maintenance planning.

Despite these promising results, there are several limitations and areas for further research:

The MRF rule is potentially applicable to any unstructured data, ranging from earthquake measurements to images and videos, suggesting broad utility beyond the current application. However, further validation with different datasets, model architectures, and application scenarios is required to ensure its general applicability. While we have shown that the MRF rule can effectively predict underfitting, its ability to reliably predict overfitting requires further investigation. Currently, the MRF rule can be used effectively only if the largest object size of interest can be determined with reasonable effort. This process could be automated by using a pyramidal feature hierarchy to automatically determine the optimal MRF above which the results no longer improve.

VADER must be validated on additional bridges to determine the influence of bridge-specific parameters such as stiffness, damping, and ballast thickness on detection performance. As VADER was initially developed for BWIM systems, which are most attractive for time-limited measurements, the results for long-term measurement campaigns were not analyzed. In particular, the influence of weather on the fundamental frequency of the bridge \citep{LorenzenDiss} and sensor drift could reduce the performance of VADER \citep{geron}.

Additionally, since VADER is trained using conventional supervised learning, labeled data for axle positions is required. However, our findings indicate that the training dataset does not need to be representative of all train types, allowing VADER to replace failed axle detectors and potentially be trained using data from individual trains equipped with DGPS. Nevertheless, because labels are never perfect and systematic labeling errors can occur with other data sources, further investigations are necessary, particularly to confirm the feasibility of the DGPS scenario with real measurement data. Moreover, only two scenarios were analyzed in our study. To further simplify the deployment of VADER, future studies should explore the feasibility of training the model with simulated data as a third scenario.

It is also unclear whether the MRF rule and VADER work effectively with other sampling rates. Therefore, the MRF rule and correspondingly optimized VADER models need to be tested at different sampling rates.

In conclusion, \(\text{VADER}_\text{raw}\) not only enhances real-time axle detection performance but also paves the way for future innovations in sensor-based monitoring systems. By potentially integrating joint signal evaluations or combining CNN-based data transformation with transformer-based classification, the VADER approach could be further refined to achieve even greater accuracy and efficiency. These advancements hold the promise of significantly improving the safety and reliability of railway infrastructure monitoring, ultimately contributing to enhanced transportation safety and more efficient maintenance strategies.

%aktualisieren
\section*{Author Contributions}
\label{S:8}
\textbf{Henrik Riedel:} Conceptualization, investigation, methodology, software, validation, visualization and writing - original draft. \textbf{Steven Robert Lorenzen:} Funding, supervision and writing - review \& editing \textbf{Clemens Hübler:} Resources, supervision and writing - review \& editing.

\section*{Acknowledgments}
\label{S:9}
The research project ZEKISS (www.zekiss.de) is carried out in collaboration with the German railway company DB Netz AG, the Wölfel Engineering GmbH and the GMG Ingenieurgesellschaft mbH. It is funded by the mFund (mFund, 2020) promoted by the The Federal Ministry of Transport and Digital Infrastructure.

The research project DEEB-INFRA (www.deeb-infra.de) is carried out in collaboration with the the sub company DB Campus from the Deutschen Bahn AG, the AIT GmbH, the Revotec zt GmbH and the iSEA Tec GmbH. It is funded by the mFund (mFund, 2020) promoted by the The Federal Ministry of Transport and Digital Infrastructure.

\begin{figure}[H]
\includegraphics[trim =55 190 120 190, clip, width=\linewidth]{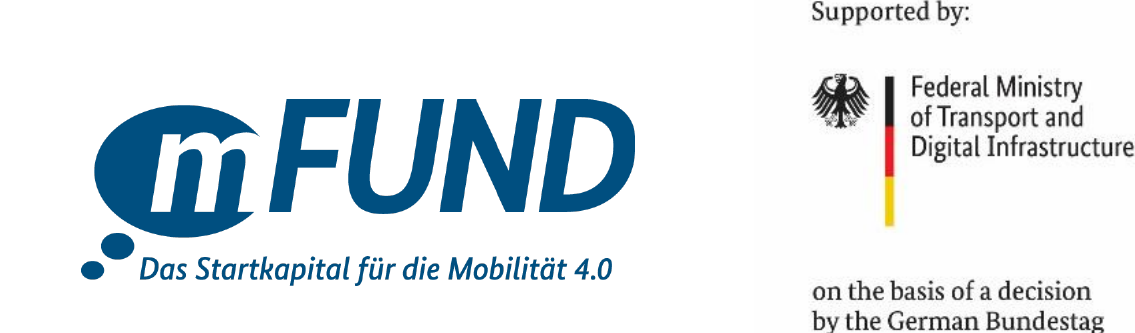}
\end{figure}

\section*{Data and Source Code}
The data \citep{Data_VADer} as well as the source code \citep{Github_VADer} used in this paper is published and contains:
\begin{enumerate}
    \item All measurement data
    \item Matlab code to label data and save as text files
    \item Python code for transformation, training, evaluation and plotting.
\end{enumerate}

%\end{linenumbers}

%% The Appendices part is started with the command \appendix;
%% appendix sections are then done as normal sections
%% \appendix

\bibliographystyle{elsarticle-num-names} 
\bibliography{References}

\end{document}